\documentclass{article}

\PassOptionsToPackage{numbers}{natbib}
\usepackage[main, final]{neurips_2026}
\usepackage[utf8]{inputenc} %
\usepackage[T1]{fontenc}    %
\usepackage{hyperref}       %
\usepackage{url}            %
\usepackage{booktabs}       %
\usepackage{amsfonts}       %
\usepackage{nicefrac}       %
\usepackage{microtype}      %
\usepackage{xcolor}         %
\usepackage{graphicx}
\usepackage{subcaption}
\usepackage{enumitem}
\usepackage{algorithm}
\usepackage[noend]{algorithmic}
\usepackage{wrapfig}
\usepackage{amsmath}
\usepackage{makecell} %
\title{On Neural Scaling Laws for Weather \\ Emulation through Continual Training}

\author{%
  Shashank Subramanian\\
  Lawrence Berkeley National Laboratory\\
  \texttt{shashanksubramanian@lbl.gov}
  \And
  Alexander Kiefer\\
  Oak Ridge National Laboratory\\
  \texttt{kiefera@ornl.gov}
  \And
  Arnur Nigmetov\\
  Lawrence Berkeley National Laboratory\\
  \texttt{anigmetov@lbl.gov}
  \And
  Amir Gholami\\
  ICSI and UC Berkeley\\
  \texttt{amirgh@berkeley.edu}
  \And
  Dmitriy Morozov\\
  Lawrence Berkeley National Laboratory\\
  \texttt{dmorozov@lbl.gov}
  \And
  Michael W. Mahoney\\
  LBNL, ICSI, and UC Berkeley\\
  \texttt{mmahoney@stat.berkeley.edu}
}

\DeclareRobustCommand{\sci}[1]{\mbox{\normalfont #1}}

\newcommand{\AR}{{\fontfamily{cmtt}\bfseries AR}}
\newcommand{\AMSE}{{\fontfamily{cmtt}\bfseries AMSE}}
\newcommand{\MSE}{{\fontfamily{cmtt}\bfseries MSE}}
\newcommand{\swin}{{\fontfamily{cmtt}\bfseries Swin}}
\newcommand{\fsdp}{{\fontfamily{cmtt}\bfseries FSDP}}

\usepackage[normalem]{ulem}

\begin{document}

\maketitle

\begin{abstract}
Neural scaling laws, which in some domains can predict the performance of large neural networks as a function of model, data, and compute scale, are the cornerstone of building foundation models in Natural Language Processing and Computer Vision.
We study neural scaling in Scientific Machine Learning, focusing on models for weather forecasting.
To analyze scaling behavior in as simple a setting as possible, we adopt a minimal, scalable, general-purpose {\swin} Transformer architecture, and we use continual training with constant learning rates and periodic cooldowns as an efficient training strategy.
We show that models trained in this minimalist way follow predictable scaling trends and can outperform standard cosine learning rate schedules.
Cooldown phases can be repurposed to improve downstream performance, e.g., enabling accurate multi-step rollouts over longer forecast horizons or sharper predictions through spectral loss adjustments.
We also systematically explore a wide range of model and dataset sizes under various compute budgets to construct IsoFLOP curves, and we identify compute-optimal training regimes.
Extrapolating these trends to larger scales highlights potential performance limits, demonstrating that neural scaling can serve as an important diagnostic for efficient resource allocation.
We open-source our \href{https://github.com/ShashankSubramanian/neural-scaling-weather}{code}
 for reproducibility.
\end{abstract}

\section{Introduction}
\label{sec:intro}

Machine learning models, and deep learning models in particular, have demonstrated great success in scientific problems, including emulating atmospheric physics for weather forecasting.
In recent years, several data-driven models \citep{lam2023graphcast,bonev2025fourcastnet3geometricapproach,bodnar2024foundationmodelearth,bi2022panguweather3dhighresolutionmodel,ben2023rise,price2023gencast,alet2025skillfuljointprobabilisticweather}  have matched or surpassed the accuracy of gold-standard Numerical Weather Prediction (NWP) systems that generate high-fidelity forecasts by solving complex, multiscale fluid-dynamics equations.
These data-driven models can generate weather forecasts orders of magnitude faster than classical NWP models using substantially lower resources, often only a single GPU, in inference.
The training costs of these deep learning weather emulators are rapidly increasing, as researchers explore a wide range of architectural choices, loss functions, and training methodologies in pursuit of state-of-the-art forecasting performance.
These costs are further amplified by the growing scale of models, reaching $O(100)$ billion parameters, as well as increasing data resolutions, e.g., through smaller patch sizes in Transformer-based architectures \citep{hatanpaa2025aerisargonneearthsystems,wang2024orbitoakridgebase,xwang2025}.
Similar scaling challenges arise in areas such as Natural Language Processing (NLP) and Computer Vision (CV).
However, scientific data present additional unique challenges, perhaps most notably their spatiotemporal structure.
The proliferation of increasingly-complex architectures, in particular in Scientific Machine Learning (SciML), raises important questions:
Are we learning about science or merely about the idiosyncrasies of ever more elaborate (domain-motivated) neural architecture designs?
What is the simplest possible architecture to obtain scaling that is good for downstream SciML applications?

These questions are timely since, in many areas of machine learning, progress has ultimately been driven not by increased architecture complexity, but by scale---more data, larger models, and more compute---a phenomenon sometimes referred to as the \emph{bitter lesson}.
Understanding how performance scales with these quantities therefore requires studying simple, general-purpose models under controlled scaling regimes.
Without such a foundation, it becomes difficult to disentangle genuine scaling behavior from artifacts introduced by specialized architectural choices.

\begin{wrapfigure}{r}{0.5\textwidth}  %
  \centering
  \includegraphics[width=0.48\textwidth]{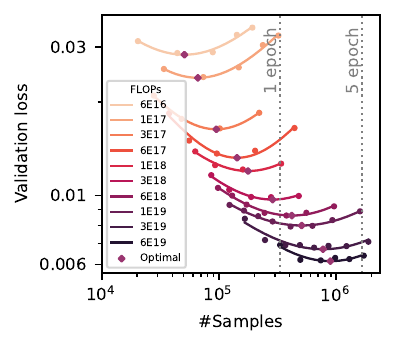}
 \caption{\textbf{Neural scaling for weather emulation.} {We pre-train several models using continual training (constant learning rates with periodic cooldowns; see \S\ref{sec:methods}), and we identify compute-optimal regimes to train the neural emulator so that neither data nor model size saturate at different compute budgets. At each FLOP budget, several model sizes (up to 457M) are trained to different dataset sizes to form IsoFLOPs that demonstrate the tradeoff between model and data size.
    Unlike NLP, these models are trained for multiple epochs (indicated by vertical dotted lines), causing samples to be revisited after the first epoch and effectively be treated as pseudo-samples.
    We fit parabolas to each IsoFLOP, and we track the compute-optimal model at each budget. }
    }
  \label{fig:scaling-data}
\end{wrapfigure}
In domains such as NLP, this challenge has been addressed through the development of \emph{neural scaling laws} \citep{hestness2017deeplearningscalingpredictable,kaplan2020scalinglawsneurallanguage,hoffmann2022trainingcomputeoptimallargelanguage}, a set of empirical patterns that guide practitioners in scaling data quantity, model size, and the amount of compute, so that none of them saturate.
When this happens, the model loss follows a power-law relationship as a function of the size of the model, of the dataset, and of the computational resources available for training.
Within NLP, these trends can span more than seven orders of magnitude \citep{hoffmann2022trainingcomputeoptimallargelanguage,grattafiori2024llama3herdmodels}.
The study of neural scaling laws is important because it builds confidence and trust that model performance will improve with increased scale.
This predictive aspect is valuable because it makes neural scaling laws not only an analytical method but also actionable for planning efficient resource allocation.

In the existing literature, there has been limited effort in understanding the neural scaling behavior in SciML, in particular for deep learning models for weather forecasting.
Existing studies do not explicitly investigate the joint relationship between model and data size, as a function of available computational budget, and/or their analyses remain incomplete.
For example, \citet{bodnar2024foundationmodelearth} report predictable scaling with model size; but their study does not perform compute-optimal analyses, relies on scaling with GPU time rather than FLOPs, and incorporates data outside the traditional weather-forecasting domain, such as pollutants and sea state, which makes interpreting the scaling behavior harder.
In contrast, \citet{karlbauer2024comparingcontrastingdeeplearning} suggest that performance stalls, for neural operators and related architectures, at even small model scales, although it should be noted that this conclusion is drawn from experiments at coarser spatial resolution than we consider.
More broadly, prior works stop short of identifying compute-optimal combinations of model size and data volume at each scale, making it difficult to disentangle true scaling laws from artifacts of architectural, data, or compute~choices.

In this work, we aim to systematically characterize the neural scaling relationships governing data-driven weather forecasting models.
Our main contributions are:
\begin{enumerate}
[nosep]
    \item
    \textbf{Minimalist transformer architecture for neural scaling.}
Rather than designing specialized architectures for weather forecasting, we intentionally adopt a minimalist and widely used backbone, the {\swin} Transformer \citep{liu2021swintransformerhierarchicalvision}, with minimal domain-specific architectural modifications or custom loss functions, during pre-training.
This design choice follows the principle that understanding scaling behavior requires reducing architectural confounders.
Our goal is therefore not to engineer the most specialized weather model, but to understand how far simple architectures can be pushed through scale.
    To enable such experiments across a wide range of scales, we further implement $2D$ spatial parallelism along with data parallelism, with the former being crucial for high-resolution inputs.
    \item
    \textbf{Continual training for weather emulators.}
    Inspired by \citet{hagele2024scalinglawscomputeoptimaltraining}, we apply learning rate (LR) cooldowns after a period of continual training with a constant LR.
    Here, ``continual training'' refers to the ability to continue optimizing a model across different compute budgets from a checkpoint without restarting from scratch, rather than the standard multi-task continual learning setting.
    We show that this fixed LR followed by a cooldown protocol can outperform
    the standard cosine schedule commonly used in pre-training weather emulators \citep{bodnar2024foundationmodelearth,lam2023graphcast} (see Fig. \ref{fig:lr-cooldown}).
    This approach enables efficient exploration of a wide range of compute budgets without retraining models from scratch, by periodically cooling down the LR to match the desired budget.
    We show that even a cooldown period as short as 5\% of the total training iterations is sufficient to achieve consistent performance (see Fig. \ref{fig:cooldown}).
    \item
    \textbf{Repurposing cooldowns for better downstream performance.}
        We demonstrate that the short cooldown period can serve a dual purpose: completing the pre-training LR schedule and adapting the model to downstream objectives, similar to task-specific fine-tuning in weather modeling.
    For example, we can perform multi-step rollouts in training during cooldown to achieve higher accuracy over longer forecast horizons (see Fig. \ref{fig:cooldown}, \ref{fig:metrics}).
    We may spectrally adjust the loss function (Fig. \ref{fig:spec}) to allow for sharper forecasts to capture critical high-resolution features.
        This allows us to disentangle pre-training scaling studies from downstream adaptation, reusing the same pre-trained checkpoints for different loss functions and rollout objectives without repeating expensive pre-training.
    \item
    \textbf{Neural scaling to identify compute-optimal training regimes.}
    We construct scaling laws by pre-training several models (up to $457M$ parameters) using compute budgets from \sci{6E16} to \sci{6E19} FLOPs on the $0.25^\circ$ ERA5 \citep{hersbach2020era5} dataset ($O(300,000)$ samples at an hourly temporal resolution).
    For each budget, different models are cooled down to different iterations that give rise to IsoFLOP curves (see Fig. \ref{fig:scaling-data}, \ref{fig:scaling-model}), a set of training configurations (model sizes and training iterations) that achieve a fixed total training FLOPs.
    Across these budgets, we observe clear compute-optimal scaling behavior, where each budget admits an optimal combination of model size and effective dataset size.
    To probe the limits of these trends, we extrapolate the scaling laws to \sci{3E20} and \sci{2.25E21} FLOPs and train $385$M and $1.24$B parameter models, respectively. Both runs show signs of saturation before reaching their projected losses.
This suggests that scaling in this regime may become limited by data size and spatiotemporal resolution, highlighting the importance of neural scaling analyses for diagnosing progress.
\end{enumerate}

\section{Related Works}
\label{sec:related_works}
\textbf{Neural weather emulators.}
Models such as FourCastNet \citep{kurth2023fourcastnet} approached the accuracy of NWP in medium-range weather forecasting (10--14 days), while being orders of magnitude faster.
Newer models \citep{bi2022panguweather3dhighresolutionmodel,lam2023graphcast,bodnar2024foundationmodelearth,bonev2025fourcastnet3geometricapproach,price2023gencast} have surpassed the accuracy of NWP, while retaining the speedups.
The architecture landscape of these models includes Neural Operators \citep{kurth2023fourcastnet,bonev2025fourcastnet3geometricapproach}, graph-based models \citep{lam2023graphcast,price2023gencast}, and Transformer-based models \citep{chen2023fengwupushingskillfulglobal,nguyen2024scalingtransformerneuralnetworks,bi2022panguweather3dhighresolutionmodel,willard2024analyzingexploringtrainingrecipes}.
In addition to a variety of architectures, a variety of loss functions, including standard MSE, autoregressive rollout MSE \citep{lam2023graphcast,brandstetter2023messagepassingneuralpde,bodnar2024foundationmodelearth}, and spectral-based losses \citep{subich2025fixingdoublepenaltydatadriven},  have been employed, depending on the downstream objective.
Despite this progress, with a focus on architectural design and accuracy, there has been relatively little effort in systematic analysis of how model design, data scale, and compute budget interact in weather emulators, which are central goals here.

\textbf{Neural scaling laws in large ML models.}
Other domains, most notably NLP, have established the utility of neural scaling laws for guiding model development.
\citet{kaplan2020scalinglawsneurallanguage} demonstrated that language model loss scales predictably as these axes increase, enabling researchers to estimate performance for larger models before training.
\citet{hoffmann2022trainingcomputeoptimallargelanguage} then formalized compute-optimal training (``Chinchilla'' scaling), showing how to better balance model parameters and training data.
Recent efforts have refined these laws, extended them to new architectures, and explored their theoretical foundations \citep{grattafiori2024llama3herdmodels,Bahri_2024,deepseekai2024deepseekllmscalingopensource,krajewski2024scalinglawsfinegrainedmixture,abnar2025parametersvsflopsscaling}.
Finally,
\citet{hagele2024scalinglawscomputeoptimaltraining} show that LR scheduler changes that allow for continual training make the cost and design of neural scaling experiments more manageable.
Our methodology is inspired by~this.

\textbf{Neural scaling laws in large SciML models.}
There are also emerging efforts to apply scaling law concepts to SciML models, beyond NLP.
For example, universal models of atomic systems have been developed with empirical scaling laws to guide capacity and data allocation across compute budgets, illustrating how scaling insights can transfer to scientific domains \citep{wood2025umafamilyuniversalmodels,wadell2025foundationmodelsdiscoveryexploration}.
Another example is Evo \citep{evo}, a foundation model that
predicts the function and structure of biological sequences and explores the design space through neural scaling laws.
Comprehensive neural scaling studies remain scarce in weather forecasting, and existing analyses do not identify compute-optimal regimes and exhibit mixed findings \citep{bodnar2024foundationmodelearth,karlbauer2024comparingcontrastingdeeplearning}.
Today, large models (billions of parameters) \citep{hatanpaa2025aerisargonneearthsystems,wang2024orbitoakridgebase} have already been trained, emphasizing the need to understand this design space.
To the best of our knowledge, this is the first work (alongside concurrent work by \citet{yu2026scaling}) to systematically study neural scaling laws in weather forecasting. \citet{yu2026scaling} scale up to \sci{1E18} FLOPs for a few existing models in the literature with sub-epoch training, and they demonstrate how some models show flavors of compute optimal scaling, whilst others do not.
The main advantage of our work is its emphasis on (i) continuous training with cooldowns for efficient neural scaling and downstream alignment, (ii) scaling to over an order of magnitude higher compute (\sci{6E19} FLOPs) via spatial parallelism to address memory constraints, (iii) demonstrating clearer compute-optimal regimes (even with multiple epochs) through a minimalist Transformer backbone that matches state-of-the-art performance under the WeatherBench metrics \citep{rasp2023weatherbench}, and (iv) probing scaling limits by extending compute to O(\sci{2E21}) FLOPs.
\section{Overview of Methods}
\label{sec:methods}
Neural scaling involves pre-training a series of models across a practical range of compute budgets.
At each compute budget, several model sizes are trained to different numbers of data samples in order to remain on an IsoFLOP.
\begin{wrapfigure}{r}{0.42\textwidth}  %
  \centering
      \includegraphics[width=\linewidth]{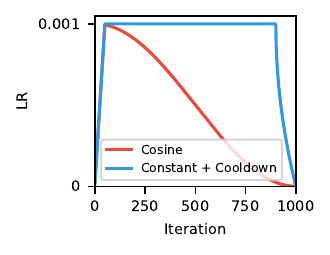}
    \caption{\textbf{LR schedules.} Cosine schedule follows a half-cosine decay after warmup; the constant$+$cooldown has a constant LR after warmup, but cools down rapidly to 0 at the end.}
    \label{fig:lr}
\end{wrapfigure}
The classical approach, based on the ``Chinchilla'' scaling laws \citep{hoffmann2022trainingcomputeoptimallargelanguage}, involves training models \emph{from scratch} for each compute budget and model size.
The need to train models from scratch is due to the reliance on the cosine LR scheduler.
\citet{hoffmann2022trainingcomputeoptimallargelanguage} showed that the final performance of any pre-trained model is optimal when the cosine LR decay length matches the total training duration.
This LR scheduler is also popular in other domains, including weather forecasting \citep{kurth2023fourcastnet,bodnar2024foundationmodelearth}. Several frontier NLP models as well as other SciML models have explored neural scaling laws in this manner \citep{grattafiori2024llama3herdmodels,deepseekai2024deepseekllmscalingopensource}.

\noindent \textbf{Scaling with periodic cooldowns.}
While the above approach is popular, it is expensive.
It would be more beneficial to change the LR scheduler to a constant schedule, followed by a rapid cooldown phase to zero LR \citep{hagele2024scalinglawscomputeoptimaltraining}.
If one could show that the performance of this constant$+$cooldown LR scheduler matches that of the cosine decay scheduler, then this would allow for training the series of models \emph{just once}, considerably reducing the computational expense of neural scaling.
We record the loss at a given compute budget by applying a LR cooldown to zero in the final pre-training iterations.
To reach larger budgets, we simply resume from the checkpoint prior to the cooldown, continue training with the original LR, and then cool down at the next target budget.

\subsection{Weather forecasting problem formulation}

Our goal is to model (emulate) global atmospheric dynamics using a learned neural network by predicting the temporal evolution of the state $\mathbf{u}(\mathbf{x},t)$, with $\mathbf{x} \in \mathbb{R}^2$ and $t \in [0, \infty)$.
Our training dataset consists of discrete snapshots $\mathbf{u}_n$ sampled at regular time steps with interval $\Delta t$ with
$\mathbf{u}_n \in \mathbb{R}^{H \times W \times C}$. Here, $H \times W$ is the height and width of the discretized latitude-longitude projection $\mathbf{u}$; $C$ is the number of state variables (as wind velocities, temperature, and others;
see \S\ref{app:era5_metrics}, with vertical dimension treated as channels).
The temporal evolution is modeled as:
\begin{equation}
    \mathbf{u}_{n+1} = \mathcal{F}_\theta(\mathbf{u}_n),
\end{equation}
where $\mathcal{F}_\theta$ is a neural network parameterized with weights $\theta$.
We consider a standard mean-squared error ({\MSE}) loss between the predicted $\mathbf{u}_{n+1}$ and target $\hat{\mathbf{u}}$ at the same time step to pre-train $\mathcal{F}_\theta$.
Once trained, the model produces forecasts during inference for $N$ steps, $t_{n+i}$ with $i \in \{1, \cdots, N\}$, by autoregressively feeding each predicted state back as input to produce subsequent forecasts.
Based on prior work, we consider two post-training strategies to better align the pre-trained model with the forecasting task:
\begin{enumerate}[leftmargin=1.9em]
    \item \textbf{Adjusted {\MSE} ({\AMSE})}: Weather datasets contain time steps ($\Delta t$) that are undersampled; they are not fine enough to sufficiently resolve fine-scale dynamics. With chaotic dynamics, this causes models trained with {\MSE} to smooth high-frequency structure. To (partially) deal with this, we consider the adjusted {\MSE} loss \citep{subich2025fixingdoublepenaltydatadriven} (see \S\ref{app:era5_metrics}), which disentangles decorrelation and spectral amplitude components of the error, enabling the model to retain high-resolution features.
    \item \textbf{Autoregressive Rollouts ({\AR})}: Autoregressive inference introduces generalization error due to a distribution shift, as predicted states are used as inputs instead of samples from the training distribution \citep{brandstetter2023messagepassingneuralpde}.
To mitigate this, models are fine-tuned to autoregressively predict multiple time steps.
While {\AR} fine-tuning reduces error over longer horizons, it leads to smooth predictions due to the chaotic dynamics, causing forecasts to behave like ensemble-means \citep{price2023gencast}.
\end{enumerate}

Existing approaches typically rely on heuristic schedules with multiple tunable components, 
including the number of autoregressive steps, learning rate schedules, and fine-tuning duration, 
for both {\AR} and {\AMSE} strategies. Here, we repurpose the cooldown period to implement either strategy, 
aligning the model to the downstream task: maintaining high spectral resolution via {\AMSE}, or reducing long-horizon forecast errors via {\AR}. 
This approach allows downstream objectives to be explored through short cooldowns 
from a shared checkpoint trained with a constant learning rate, without needing to repeat pre-training for each choice.

\subsection{Architecture and Training}

We employ a standard architecture based on the Shifted Window ({\swin}) Transformer \citep{liu2021swintransformerhierarchicalvision}.
While many weather models introduce domain-specific modifications and customized components, we follow \citet{willard2024analyzingexploringtrainingrecipes}, who demonstrate that standard transformer backbones can be highly competitive when trained well.
Our input $\mathbf{u}_n$ is projected to the latent space $\mathbf{X} \in \mathbb{R}^{H/p \times W/p \times E}$ using a standard patch embedding layer with patch size $p$ and embedding dimensions $E$.
We add a simple coordinate-based positional encoding using coordinates $(x, y, z)$
(derived from spherical coordinates) and normalized time step $t \in \mathbb{R}$ for each pixel, projected to the latent space $\mathbf{PE} \in \mathbb{R}^{H/p \times W/p \times E}$ via an additional patch embedding layer. The positional encoding is added to the latent feature $\mathbf{X}$ before being passed to the {\swin} Transformer blocks. While alternative positional encoding strategies exist, we avoid learnable encodings, which would introduce a large number of parameters at high input resolutions (especially for small patch sizes $p$) and risk over-parameterizing this component relative to the rest of the model.
We find this positional encoding to be sufficient in our experiments.
For the {\swin} blocks, we remove components such as relative positional bias and other non-essential modifications, including the hierarchical {\swin} structure (e.g., patch merging and downsampling), and we retain a uniform Transformer block design consisting of windowed multi-head self-attention (\text{W-MHSA}) and a multi-layer perceptron (MLP) with pre-RMSNorm and residual connections:
\begin{align*}
    \tilde{\mathbf{X}} &= \text{RMSNorm}(\mathbf{X}), \quad \mathbf{Y} = \text{W-MHSA}(\tilde{\mathbf{X}}) + {\mathbf{X}}, \\
    \tilde{\mathbf{Y}} &= \text{RMSNorm}(\mathbf{Y}), \quad \mathbf{O} = \text{MLP}(\tilde{\mathbf{Y}}) + {\mathbf{Y}},
\end{align*}
where the \text{W-MHSA} operation partitions the latent patch grid into $2$D windows and computes standard dot-product self-attention locally.
We use QK-normalization for training stability \citep{henry2020querykeynormalizationtransformers}.
Every alternate block shifts the windows to enable cross-window interactions, implemented as a rolling operation with attention masks. We modify the mask to preserve the left-right periodicity of the Earth system. The final layer is a reverse patch embedding~layer.

\noindent \textbf{Distributed training for neural scaling.}
As model, data, and input sizes increase, standard data parallelism becomes insufficient, with per-GPU batch sizes dropping to one.
While NLP models address this with model parallelism, SciML domains such as weather face different challenges.
High-resolution inputs and small patch sizes make these models heavily constrained by intermediate activation memory rather than weights, a limitation that is amplified during autoregressive rollout ({\AR}) in training. Standard techniques such as Fully-Sharded Data Parallelism ({\fsdp}) do little to alleviate this memory pressure.
Moreover, I/O pressure increases due to high-resolution data.
It is essential to build efficient distributed infrastructure to train across a wide range of scales.
Towards this end, we implement spatial parallelism through domain decomposition, along with data parallelism.
We employ a 3D mesh of GPUs $n_1 \times n_2 \times n_d$, where the last
axis handles data parallelism and the first two partition the input $H \times W$.
Since activation memory scales as $O(BHWE)$, with batch size $B$ and embedding dimension $E$, this partitioning effectively reduces this memory to $O(BHWE/n_d n_1 n_2)$. The shifting windows in {\swin} require additional communication that we implement with a custom distributed roll in the forward and backward pass (for details, see \S\ref{app:hp}).

\subsection{Empirical Analysis Setup}
\noindent \textbf{Data.}
Our models are trained on the ERA5 \citep{hersbach2020era5} dataset that consists of around $350,000$ samples of the Earth's atmospheric state from 1979 to 2022 at hourly time steps and spatial resolution of $0.25^\circ$. While ERA5 consists of hundreds of state (physical) variables, we use a 71 variable subset following \citet{willard2024analyzingexploringtrainingrecipes} (see \S\ref{app:era5_metrics}).
When projected on a latitude-longitude grid, each sample is a $71 \times 721 \times 1440$ tensor with $721 \times 1440$ representing the spatial grid. We use the years 1979 to 2016 as training data, 2017 as validation, and 2020 for testing. We train our models to predict a time step of 6 hours.

\noindent \textbf{Loss and Metrics.}
Our pre-training loss is {\MSE}. We avoid domain-specific weighting strategies that are variable dependent in the loss to maintain a simple objective for the neural scaling analysis.
Our inputs are normalized with the per-variable mean and standard deviation derived from the training dataset.
For validating our results, we use the area-weighted Root Mean-Squared Error (RMSE) and Power Spectral Density (PSD) prediction and target for each variable.
We track each metric as a function of lead time up to 10 days, corresponding to the medium-range forecasting regime, and average the results over multiple initial conditions sampled across the evaluation period. We follow this evaluation protocol on initial conditions sampled regularly for the testing year 2020 (every 12 hours, 732 initial conditions) following the WeatherBench protocol \citep{rasp2023weatherbench}.
We also retrieve the benchmark model predictions from Google Cloud Storage (using {\fontfamily{cmtt}\bfseries gcsfs}), following \citet{rasp2023weatherbench}.
While RMSE provides a measure of average error over the globe, the PSD is commonly used to assess the effective resolution of the predictions and to quantify how well high-frequency, fine-scale features are captured.
See \S\ref{app:era5_metrics} for the definitions.

\noindent \textbf{Training.}
We train all our models with a batch size of 16.
Through tuning experiments, we select a peak LR of \sci{5E-4} for models larger than $100M$ parameters. Similar to NLP \citep{kaplan2020scalinglawsneurallanguage}, we find smaller models can tolerate higher peak LRs, with the optimal LR approximately inversely proportional to the square root of the parameter count.
We train our models with the AdamW optimizer with a small weight decay of \sci{1E-4} and gradient norms clipped at $1$. We find that this, with QK-normalization, is useful for training stability at the higher parameter counts.
We use the constant$+$cooldown LR scheduler of the scaling experiments and cosine for initial comparisons. Both schedulers use 500 warmup iterations.
We use mixed precision with \texttt{float16} for training.
See Tab. \ref{tab:configs} for all hyperparameter choices.

\section{Main Results}
\label{sec:results}

We demonstrate the following results:
\textbf{(R1)} Models can be continuously trained with periodic cooldowns using simple Transformer backbones;
\textbf{(R2)} Cooldown periods can be repurposed with alternate losses to align to a weather-specific downstream task; and
\textbf{(R3)}  Compute-optimal regimes, through systematic neural scaling using cooldowns, can be identified across multiple compute budgets.

\begin{figure}
    \centering
    \begin{subfigure}[b]{.45\textwidth}
        \centering
        \includegraphics[width=\textwidth]{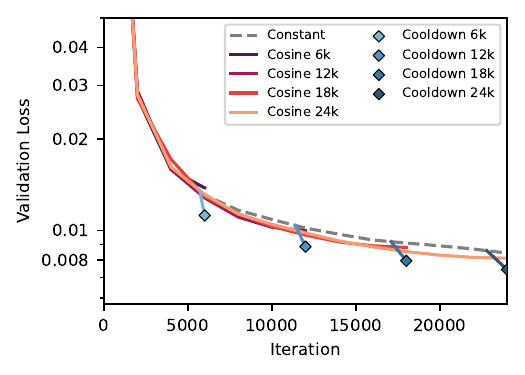}
        \caption{Cosine vs. Cooldowns}
        \label{fig:lr-cooldown}
    \end{subfigure}%
    \begin{subfigure}[b]{.45\textwidth}
        \centering
        \includegraphics[width=\textwidth]{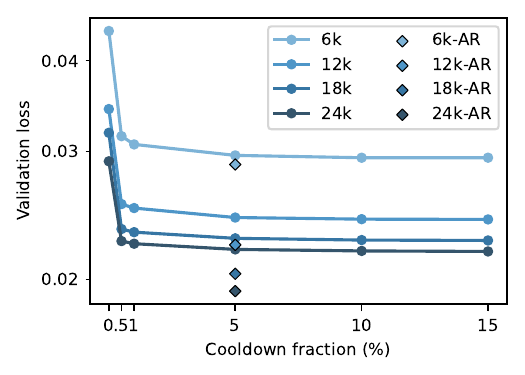}
        \caption{Cooldown fraction and alignment}
        \label{fig:cooldown}
    \end{subfigure}

    \caption{\textbf{(left) Cooldowns are more performant than cosine.} The validation loss of the model continuously trained with a constant LR and cooled down to different iteration counts (5\% cooldown) shows better losses compared to the {\swin} trained from scratch with different cosine schedulers that match the total iteration count.
    \textbf{ (right) Short cooldowns are sufficient and can help alignment.}  We show the validation loss over 36 hours (6 {\AR} steps) of prediction averaged over the validation data as a function of the fraction of total iterations used in cooldown. The loss decreases with longer cooldowns and this is true over multistep predictions. At $\sim$5\%, the gains plateau. When the cooldown is repurposed with 4-step {\AR}, we see lower errors across the longer horizon.}
    \label{fig:r12}
\end{figure}

\textbf{(R1) Cosine vs.\ Cooldowns.}  We first train a moderately sized $115M$ parameter {\swin} (see architecture hyperparameters in Tab. \ref{tab:configs}) from scratch multiple times using the cosine LR schedule, each run to a different number of iterations (batch size 16,  corresponding to 20,819 iterations per epoch).
We use the standard MSE loss with one-step prediction (predict the state $\Delta t = 6$ hours ahead).
We then replace this setup with the constant$+$cooldown LR schedule, cooling down to the same final iteration counts as the cosine runs using only 5\% of the total iterations.
We observe that the models that have been cooled down consistently show lower validation loss compared to the cosine LR (see Fig.~\ref{fig:lr-cooldown} for the validation loss curves for each run). 
Independent LR sweeps for both schedules confirm lower cooldown losses across the tested LRs and budgets (Tab.~\ref{tab:lr-sweep}), supporting its use for efficient scaling studies. We show in Fig. \ref{fig:vit}, \ref{fig:fcn} that other architectures also show similar behavior.
In order to use the cooldowns for neural scaling, we investigate what the optimal fraction of the total iterations should be used for the cooldown period.
We repeat the cooldown experiments using different cooldown fractions (from 0\% to 15\%) for each final iteration count.
We note that while the one-step prediction is important for training, the true utility of a neural emulator lies in its ability to produce reliable forecasts over longer time horizons using autoregressive inference, as described in \S\ref{sec:methods}.
Hence, we perform autoregressive rollouts during validation over multiple steps to track the time-averaged loss. Importantly, a cooldown that improves one-step loss but does not translate to longer-horizon accuracy may reduce its practical benefits.
In Fig.~\ref{fig:cooldown}, we track the validation loss over six steps (36 hours), and we find that the gains from cooldown translate to the downstream forecasting task as well.
We observe that even a small cooldown fraction (0.5\%) is able to quickly reduce the loss and also see that a cooldown of 5\% is sufficient, beyond which the gains plateau.
We use 5\% cooldowns in all experiments henceforth.

\begin{figure*}
    \centering
    \begin{subfigure}[b]{\textwidth}
        \centering
        \includegraphics[width=\textwidth]{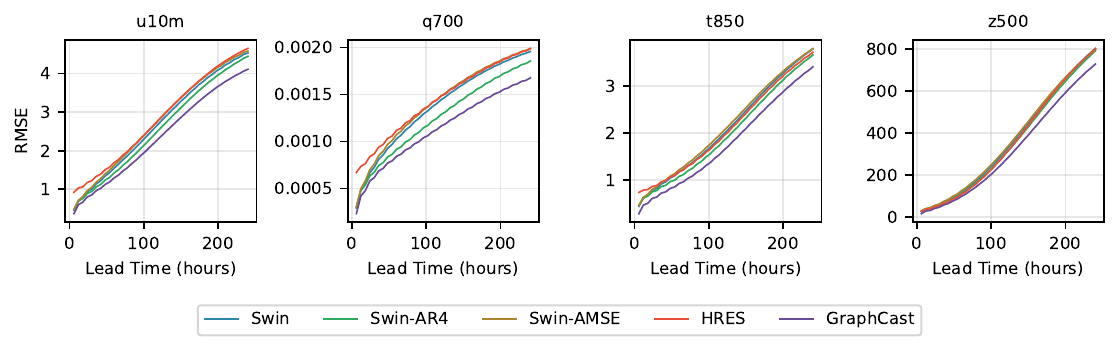}
        \caption{RMSE of different variables (labeled at the top) vs. lead time (in hours) for different models.}
        \label{fig:rmse}
    \end{subfigure}
    \hfill
    \begin{subfigure}[b]{\textwidth}
        \centering
        \includegraphics[width=\textwidth]{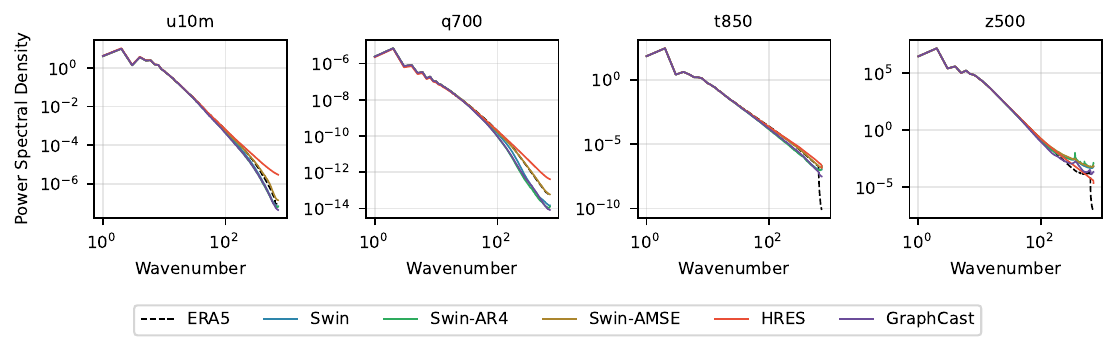}
        \caption{PSD of different variables compared to PSD of the truth (ERA5) for different models at 24 hours lead time.}
        \label{fig:spec}
    \end{subfigure}

    \caption{\textbf{Cooldowns can be used for alignment.}  When evaluated on testing year, the {\swin} cooled down at 24000 iterations is able to surpass the NWP (HRES). When {\AR} is used in cooldown ({\swin}-{\AR}4), the RMSE drops further, consistent with the use of this loss.  When {\AMSE} is used ({\swin}-{\AMSE}), the PSD retains high wavenumbers. This is easily seen in $q700$ where the {\AMSE} spectrum closely matches that of ERA5, but other models blur significantly. {\AR} contributes to more blurring in favor of reduced RMSE (visible in dissipation of power in high wavenumbers). We note that HRES models weather at a $0.1^\circ$ resolution and hence shows higher resolution.
    While this {\swin} does not beat the state-of-the-art GraphCast, a larger {\swin} obtained through optimal scaling matches its performance (as demonstrated later in Fig. \ref{fig:rmse-opt}).}
    \label{fig:metrics}
\end{figure*}

\begin{figure*}
    \centering
    \includegraphics[width=1\linewidth]{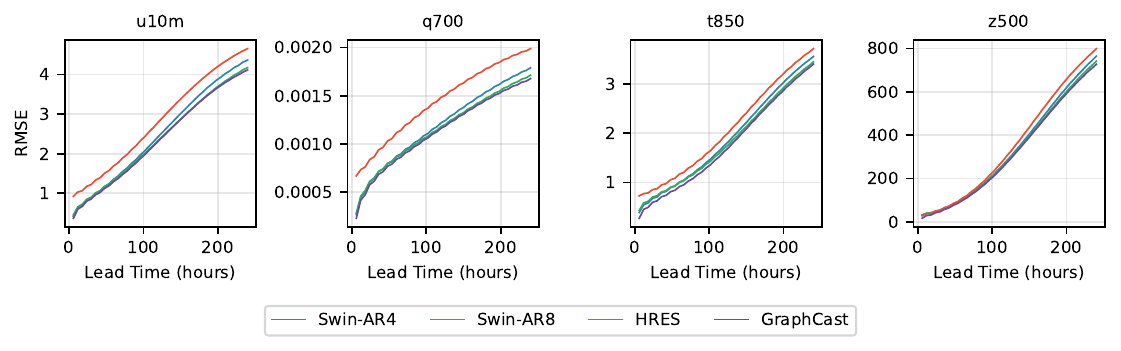}
    \caption{\textbf{RMSE performance of compute-optimal {\swin} vs baselines.} The 204M {\swin} (compute-optimal \sci{6E19} FLOPs) outperforms NWP / HRES and is on par with GraphCast when evaluated over initial conditions in 2020. The model is cooled down with 4-step {\AR}. With more steps (8-step {\AR}), the performance gap reduces at longer lead times, as expected (GraphCast uses 2--12-step {\AR}).}
    \label{fig:rmse-opt}
\end{figure*}

\textbf{(R2) Alignment with Cooldown.}
We demonstrate that the cooldown can be used to also align the model better with the downstream task of high-resolution forecasting for several time steps.
Similar to task-specific fine-tuning, this adaptation uses alternative training objectives, but is integrated into the short LR cooldown.
This is important because neural scaling can be conducted using a simple {\MSE} objective on large models and data, while the model's behavior can be adapted during the cooldown using alternative objectives. As a result, the same pre-training scaling study can support multiple downstream adaptation objectives.
We consider both autoregressive training ({\AR}) and the {\AMSE} loss for this purpose (as described in \S\ref{sec:methods}). While {\AR} aligns the model for long-term forecasting skill (through smoother forecasts like an ensemble), {\AMSE} aligns the models to produce high-resolution features.
In Fig.~\ref{fig:cooldown}, we show that if the model is cooled down with 4 steps of {\AR}, the validation loss systematically improves at all the iteration counts. In Fig.~\ref{fig:rmse}, we show the RMSE of the {\swin} predictions as a function of lead time for several variables, compared to the NWP gold standard (HRES) \citep{ifs_hres_dataset} as well as the deterministic state-of-the-art emulator GraphCast \citep{lam2023graphcast}.  We see that the {\swin} is performant (outperforming HRES) and can be systematically pushed to the GraphCast accuracy through {\AR} (note that GraphCast was fine-tuned with 2 through 12-step {\AR}).
We can also see from the PSD metric in Fig.~\ref{fig:spec} that the RMSE gains occur through smoothing that emulates an ensemble prediction.
We show later in Fig. \ref{fig:rmse-opt} that when optimally scaled, {\swin} can match GraphCast.
Similarly, using {\AMSE} during cooldown, we see that RMSE remains largely unaffected; however, the PSD plots show that the model has more power in the higher wavenumbers that this loss was designed to capture.
We emphasize that the goal here is not to achieve best possible performance, but rather to demonstrate that phases such as cooldowns can be repurposed to align model behavior, highlighting their flexibility without entangling them with the core scaling analysis.

\begin{figure*}
    \centering
    \includegraphics[width=\linewidth]{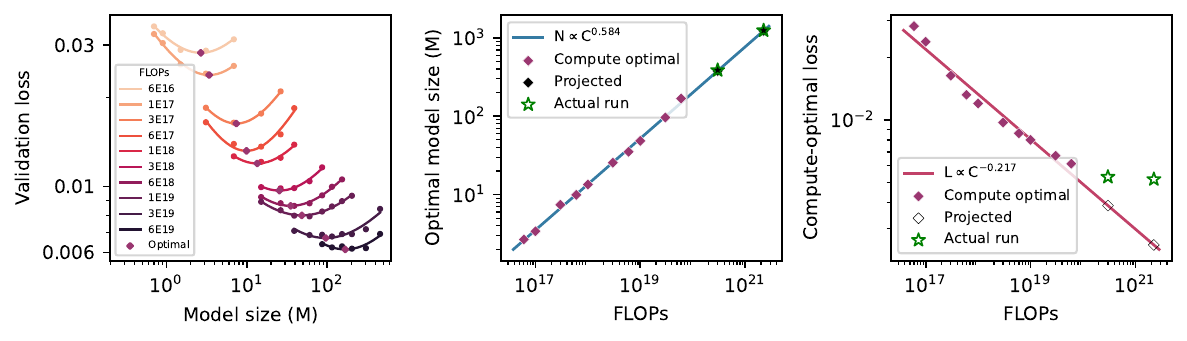}
        \caption{\textbf{Optimal model sizes as a function of compute.}
        (left) Similar to Fig. \ref{fig:scaling-data}, we show the validation loss as a function of model size for different compute budgets.
        For each budget, the different model sizes are trained to a different number of iterations to create an IsoFLOP.
        We track the minima for each IsoFLOP (through a fitted parabola).
        (middle) We fit an empirical scaling law to find the optimal model size for any FLOP budget and project to
        two FLOP budgets: \sci{3E20} and \sci{2.25E21} FLOPs to find the optimal model.
        (right) We project the loss to the projected FLOP values---the measured loss starts to saturate.}
    \label{fig:scaling-model}
\end{figure*}

\textbf{(R3) Neural Scaling.}
We train a range of model sizes from $0.7M$ to $457M$ parameters (see all hyperparameters in Tab. \ref{tab:configs}), varying more than two orders of magnitude.
Our patch-size, embedding-width, and depth ablations (Tab.~\ref{tab:patch}) motivate using patch size 4.
Width and depth are largely interchangeable away from extreme aspect ratios,
with a preference for deeper models, possibly due to our minimalist architecture and higher compute budgets.
This differs from the preference for wider models in the smaller compute range studied by \citet{yu2026scaling}.
We select three orders of magnitude of compute budgets from \sci{6E16} to \sci{6E19}.
For each budget, we estimate (see \S\ref{app:hp} for details on FLOP estimates) a range of model sizes and number of iterations to train such that each configuration lies on an IsoFLOP. Each model is trained only once and cooled down to the required iteration at each IsoFLOP. We note that several of our models train for more than one epoch.
In Fig.~\ref{fig:scaling-data}, we show the IsoFLOP curves on the loss versus number of samples scaling plot, while Fig. \ref{fig:scaling-model} presents the corresponding scaling behavior of loss versus model size. For both plots, we fit parabolas to each IsoFLOP to estimate the compute-optimal dataset size and model capacity for a given budget (see \S\ref{sec:sens} for sensitivity analyses).

\begin{figure*}
    \centering
    \includegraphics[width=1\linewidth]{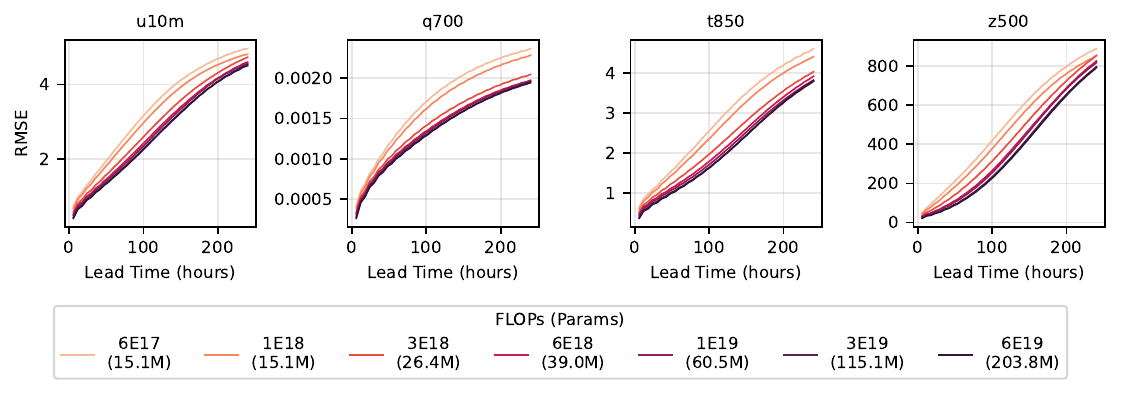}
    \caption{\textbf{RMSE performance of the closest compute-optimal {\swin} models at each compute budget.} We find that RMSE at extended forecast horizons, up to 240 hours, consistently improves with increased compute. However, at the highest compute levels, the rollout performance begins to exhibit signs of saturation.}
    \label{fig:rmse-flops}
\end{figure*}

We observe that the minimalist model \textit{(i)} scales systematically to \sci{6E19} FLOPs without saturating the loss  and \textit{(ii)} exhibits clear compute-optimal behavior where each budget affords an optimal model size and sample count (as opposed to other specialized architectures studied by \citet{yu2026scaling}). Interestingly, this trend persists at several budgets where the model is trained for multiple epochs, demonstrating the value of data reuse.
 In Fig. \ref{fig:rmse-flops} (and corresponding spectra in Fig. \ref{fig:spectra-flops}), we show that metrics maintain their gains throughout the rollout and across different compute budgets.
In Fig. \ref{fig:rmse-opt}, we show that the model that is closest to compute-optimal at \sci{6E19} FLOPs ($204M$) matches the state-of-the-art GraphCast (with other variables reported in Fig. \ref{fig:rmse-all}).
We estimate that GraphCast, while smaller ($37M$), trained longer and used close to O(\sci{1E21}) FLOPs (see \S\ref{sec:flops} for our estimation), with about 4 weeks of training \cite{lam2023graphcast} on 32 TPUv4 chips. In contrast, the $204M$ {\swin} trains with 32 NVIDIA A100s (comparable hardware generation) in about 6 hours (\sci{6E19} FLOPs) due to more efficient Transformer operations compared to (memory-bound) graph operations (and gradient checkpointing to save memory).

We then derive the following scaling: $S^\star(C) \propto C^{0.431}$ and $N^\star(C) \propto C^{0.584}$, where $C$ is the budget,
$S^\star$ is the optimal number of (pseudo) samples (with the understanding that multiple epochs are possible)
and $N^\star$ is the optimal model size (see Fig. \ref{fig:scaling-model}).
We extrapolate to two FLOP budgets: \sci{3E20} and \sci{2.25E21} FLOPs ($5\times$ and $37.5\times$ the largest fitted budget of \sci{6E19}, respectively)
to investigate how long the scaling holds.
The empirical scaling suggests training the following two models: $379M$ and $1.23B$.
We train a $385M$ model to $122K$ iterations (\sci{3E20} FLOPs) and $1.24B$ parameter model for $288K$ iterations (\sci{2.25E21} FLOPs).
We observe in Fig. \ref{fig:scaling-model} that both these models (denoted by stars) saturate before hitting their projected loss values.
The $385M$ and $1.24B$ runs correspond to
approximately 6 and 14 epochs, respectively.
The training and validation losses in Fig.~\ref{fig:loss-iter} support overfitting from repeated exposure to finite data as the most likely explanation for this saturation.
Stronger regularization reduces the train--validation gap of the $1.24B$ model, but it does not recover the extrapolated loss (Fig.~\ref{fig:loss-iter}),
further supporting this interpretation.
Thus, our scaling laws characterize a finite-data regime in which training repeatedly samples ERA5.
Consistent with observations in data-constrained language models \citep{muennighoff2023scaling},
data reuse remains useful over several epochs but offers diminishing returns.
The weaker IsoFLOP fits at higher budgets and the deviations from the loss--compute trend in both extrapolation runs suggest
limits to extending these empirical power laws beyond the fitted regime. Restricting the analysis to a single epoch avoids data repetition
but leaves too few well-resolved IsoFLOP minima for reliable fitting (see the multi-epoch analysis in \S\ref{app:data-reuse}).
This provides early evidence that scaling trends may begin to break down in this regime, strongly motivating
this kind of neural scaling analysis before moving SciML models to frontier model scales, data scales, and resolutions.

\section{Conclusions}
\label{sec:conclusions}
We systematically studied neural scaling for data-driven weather forecasting SciML models.
To do so, we used a minimalist Transformer backbone, without much domain-specific complexities, so that scaling behavior could be studied in its simplest and most interpretable form, without confounding architecture / loss function choices.
We used a continual training approach with constant LRs and periodic cooldowns, an approach which offers several advantages.
This outperforms the commonly-used cosine decay schedule in the tested settings; it enables rapid alignment of the model with downstream objectives during short cooldown phases; and, most importantly, it makes scaling law analysis practical by allowing IsoFLOP curves to be constructed without repeatedly training models from scratch.
We demonstrated this by efficiently exploring a wide range of model sizes and compute budgets, identifying compute-optimal relationships between model size, dataset size, and training compute, with neural scaling trends emerging across three orders of magnitude in compute (even with multi-epoch training).
Our extrapolation runs at larger compute budgets show signs of saturation.
These results support limited data quantity and diversity as the most likely constraints in this multi-epoch regime.
Even when following compute-optimal scaling, large models require multiple passes over the data, so simply increasing model size may not yield proportional gains.
These observations suggest that very large SciML models should be carefully analyzed and planned before scaling further.
A natural next step is to extend cooldown adaptation and scaling studies to probabilistic objectives such as CRPS, evaluating calibration and whether short cooldowns remain sufficient.
Overall, our results emphasize that understanding neural scaling behavior for a given dataset is critical before training SciML models at frontier scale, and they provide a practical framework for doing so in spatiotemporal scientific domains.

\begin{ack}
This research used the Perlmutter supercomputing resources of the National Energy Research Scientific Computing Center (NERSC), a U.S. Department of Energy Office of Science User Facility located at Lawrence Berkeley National Laboratory, operated under Contract No. DE-AC02-05CH11231.
An award of computer time was provided by the ASCR Leadership Computing Challenge (ALCC) program at NERSC under ERCAP0038267.
This work was also supported by the NERSC Science Acceleration Program (NESAP) at NERSC under ERCAP0038454, 
and supported by resources of the Oak Ridge Leadership Computing Facility at the
Oak Ridge National Laboratory, which is supported by the Advanced Scientific
Computing Research programs in the Office of Science of the U.S. Department of
Energy under Contract No. DE-AC05-00OR22725.
SS would like to thank Ankur Mahesh for helpful discussions on weather modeling and downstream metrics.
Our conclusions do not necessarily reflect the position or the policy of our sponsors, and no official endorsement should be~inferred.
\end{ack}

{
\small
\bibliography{example_paper}
\bibliographystyle{plainnat}
}
\newpage
\appendix
\onecolumn

\renewcommand{\thefigure}{A\arabic{figure}}
\setcounter{figure}{0}
\renewcommand{\tablename}{Table}   %
\renewcommand{\thetable}{A\arabic{table}}
\setcounter{table}{0}
\section{Appendix : Additional Details}
\label{sec:appendix}
\subsection{Limitations}
\label{sec:limitations}
The main limitation of this paper is its focus on deterministic training of the weather model. Given the inherently chaotic nature of weather, probabilistic estimates are often more informative for forecasting at future time points, with metrics such as continuous ranked probability scores (CRPS) and spread--skill ratios (SSR) providing better diagnostics for a model's predictive capability. Addressing this would involve training (or cooling down) models using diffusion objectives \citep{price2023gencast,Alexe2024AIFS} or CRPS-based ensemble training \citep{Alexe2024AIFS,bonev2025fourcastnet3geometricapproach}, which may exhibit different scaling behaviors; incorporating physics-informed penalties during cooldown and characterizing how upstream scaling translates to quantitative downstream metrics (such as cyclone-track error) are also left for future work. Our patch size is fixed for the scaling tests, but the internal Transformer resolution (sequence length) represents another dimension for systematic scaling studies. Another possible limitation is the focus on Transformer architectures. While this focus is intentional to demonstrate that simple architectures can scale and perform competitively, it remains important to explore other architecture types, such as FourCastNet \cite{bonev2025fourcastnet3geometricapproach}, which incorporates geometric inductive biases through spherical transforms within the neural operator framework, or graph-based approaches like GraphCast \cite{lam2023graphcast}.
Finally, to rigorously probe the implications of scaling laws and not be constrained by multi-epoch training, it is essential to extend beyond this single domain, incorporating diverse datasets from across the Earth Sciences (and beyond), and exploring cross-domain foundation model pre-training. This will allow us to evaluate scaling behavior in a broader, more complex data landscape.

\subsection{ERA5 Dataset and Metrics}
\label{app:era5_metrics}
For the current weather state $u_n \in \mathbb{R}^{71 \times 721 \times 1440}$, we closely follow existing works \citep{bonev2025fourcastnet3geometricapproach,willard2024analyzingexploringtrainingrecipes} and select the following variables: geopotential height ($z$), winds ($u$, $v$), temperature ($t$), and specific humidity ($q$) at 13 vertical pressure levels ($50$hPa, $100$hPa, $150$hPa, $200$hPa, $250$hPa, $300$hPa, $400$hPa, $500$hPa, $600$hPa, $700$hPa, $850$hPa, $925$hPa, and $1000$hPa).
We also include the following surface variables: $u10m$ ($u$ at $10$m), $v10m$ ($v$ at $10$m), $t2m$ (temperature at $2$m), surface pressure ($sp$), mean sea level pressure ($msl$), and total column water vapor ($tcwv$).
Finally, we include land-sea mask, orography, and cosine of the latitude (to convey the curvature of the earth) as static (not dependent on time) masks (not predicted).

Our metrics are area-weighted RMSE and PSD for each channel $C$ defined as follows. Given prediction $u_n$ and ground truth $\hat{u}_n$ at time step $n$,
\begin{equation*}
\mathrm{RMSE}(u_n, \hat{u}_n) = \sqrt{
\frac{1}{H \, W}
\sum_{h=1}^{H} \sum_{w=1}^{W} w_h
\big( u_{c,h,w} - \hat{u}_{c,h,w} \big)^2
},
\end{equation*}
with $w_h$ being the area weights, proportional to the cosine of the latitude, normalized to one.

The PSD \citep{WikipediaSphericalHarmonics2026,bonev2025fourcastnet3geometricapproach} of $u$ at spherical harmonic degree $\ell$ is defined as:
\begin{equation}
\mathrm{PSD}_\ell(u) = \frac{1}{2\ell + 1}\sum_{m=-\ell}^{\ell} |u_{\ell m}|^2,
\end{equation}
where $u_{\ell m}$ are the coefficients of $u$ in the spherical harmonic decomposition:
\[
u(\theta, \phi) = \sum_{\ell=0}^{\infty} \sum_{m=-\ell}^{\ell} u_{\ell m} \, Y_{\ell m}(\theta, \phi),
\]
with $Y_{\ell m}$ denoting the spherical harmonic functions of degree $\ell$ and order $m$, and $(\theta, \phi)$ the latitude and longitude coordinates.
We use {\fontfamily{cmtt}\bfseries torch-harmonics} \citep{bonev2023spherical,kurth2026library} to implement the Spherical Harmonic Transform (SHT).
We apply all SHTs in FP64 to avoid numerical issues.
  We use the SHTs for the metrics as well as the {\AMSE} loss functions, defined below.

For the {\AMSE} loss \citep{subich2025fixingdoublepenaltydatadriven}, we use the total power at each degree,
\begin{equation}
P_\ell(u) = \sum_{m=-\ell}^{\ell}|u_{\ell m}|^2
= |u_{\ell 0}|^2 + 2\sum_{m=1}^{\ell}|u_{\ell m}|^2
= (2\ell+1)\mathrm{PSD}_\ell(u).
\end{equation}
The second equality holds for real-valued fields and accounts for the negative orders omitted by the real SHT. The loss between $u$ and $\hat{u}$ is

\begin{equation}
\mathrm{AMSE}(u, \hat{u}) = \sum_\ell \Bigg[
\big( \sqrt{P_\ell(u)} - \sqrt{P_\ell(\hat{u})} \big)^2
+ 2 \, \max\big( P_\ell(u), P_\ell(\hat{u}) \big) \, \big( 1 - \mathrm{Coh}_\ell(u, \hat{u}) \big)
\Bigg],
\end{equation}
with the coherence between $u$ and $\hat{u}$ at degree $\ell$ as:
\begin{equation}
\mathrm{Coh}_\ell(u, \hat{u}) = \frac{\sum_{m=-\ell}^{\ell} \mathrm{Re} \big( u_{\ell m} \, \hat{u}_{\ell m}^* \big)}
{\sqrt{P_\ell(u) \, P_\ell(\hat{u})}}.
\end{equation}

\begin{table}
\caption{Model configurations used for the neural scaling: We run each of these models once to different final iteration counts based on the required FLOP budgets with 5\% cooldown periods at the end for the LR.}
\label{tab:configs}
\begin{center}
\begin{tabular}{cccc}
\hline
Param (M) & Embed & Depth & LR \\
\hline
0.7 & 96 & 4 & \sci{1.5E-3} \\
0.9 & 96 & 6 & \sci{1.5E-3} \\
1.5 & 128 & 6 & \sci{1.5E-3} \\
3.1      & 192  & 6  & \sci{1.5E-3} \\
6.9      & 256  & 8  & \sci{1.5E-3} \\
15.1     & 384  & 8  & \sci{1.5E-3} \\
26.4     & 512  & 8  & \sci{1.2E-3} \\
39.0     & 512  & 12 & \sci{1E-3} \\
60.5     & 640  & 12 & \sci{8E-4} \\
86.8     & 768  & 12 & \sci{6E-4} \\
115.1    & 768  & 16 & \sci{5E-4} \\
153.5    & 1024 & 12 & \sci{5E-4} \\
203.8    & 1024 & 16 & \sci{5E-4} \\
304.5    & 1024 & 24 & \sci{5E-4} \\
456.7    & 1536 & 16 & \sci{5E-4} \\
\hline
\end{tabular}
\end{center}
\end{table}

\subsection{Model hyperparameters and training details}
\label{app:hp}
We train the {\swin} models (each with global batch size 16) summarized in Tab. \ref{tab:configs}.
We fix our patch size to 4 based on patch size ablations (see Tab. \ref{tab:patch})
We also fix the head dimension to $64$ through minimal hyperparameter tuning experiments and the window size to $9 \times 18$ based on \citet{willard2024analyzingexploringtrainingrecipes}. We observe minimal accuracy gains with larger windows and smaller ones begin to degrade the accuracy. For the feedforward (MLP) layers, we use a hidden size to embedding dimension ratio of $4$ for all models.
For the $1.24B$ {\swin}, we use an embedding dimension of $1792$ and depth $32$.
For the cooldowns, we use a 1-sqrt() cooldown function based on \citet{hagele2024scalinglawscomputeoptimaltraining}.

\medskip\noindent\textbf{LR tuning.}
\label{app:lr-tuning}
For cosine vs. cooldowns, we independently sweep the peak LR over $\{\sci{1E-4}, \sci{5E-4}, \sci{1E-3}\}$ for the $115M$ model, with global batch size 16 and the same one-step MSE training objective. We compare schedules at matching total iteration budgets of 12K, 18K, and 24K, including the final 5\% cooldown. We report the validation losses in Table~\ref{tab:lr-sweep}. Constant$+$cooldown achieves lower loss at each tested LR and budget. Hence, this is a valid (and better performing) strategy for continual training to carry out cheaper neural scaling experiments in our setting.
For the scaling experiments, we select a peak LR of \sci{5E-4} for models larger than $100M$ parameters. We find that smaller models tolerate higher LRs, up to \sci{1.5E-3}, similar to observations in NLP \citep{kaplan2020scalinglawsneurallanguage}, whereas the larger models eventually diverge. In our tuning experiments with moderately sized models, the selected LR is approximately inversely proportional to the square root of the parameter count. The model-specific LRs are listed in Tab.~\ref{tab:configs}. For the $1.24B$ model, we did not test a higher LR because larger LRs had eventually diverged for the earlier model runs and tuning at this scale was prohibitively expensive.

\begin{table}[t]
\centering
\caption{Independent LR sweeps for the $115M$ {\swin}. Validation loss at matched total iteration budgets (including cooldown); lower is better. Both schedules use global batch size 16. Constant$+$cooldown uses the final 5\% of iterations for cooldown.}
\label{tab:lr-sweep}
{
\begin{tabular}{rlrrr}
\toprule
Iterations & Schedule & \sci{1E-4} & \sci{5E-4} & \sci{1E-3} \\
\midrule
12K & Cosine & 0.026 & 0.010 & 0.009 \\
    & Constant$+$cooldown & 0.019 & 0.0089 & 0.0082 \\
\midrule
18K & Cosine & 0.020 & 0.0088 & 0.0079 \\
    & Constant$+$cooldown & 0.014 & 0.0080 & 0.0074 \\
\midrule
24K & Cosine & 0.017 & 0.0081 & 0.0074 \\
    & Constant$+$cooldown & 0.012 & 0.0075 & 0.0071 \\
\bottomrule
\end{tabular}
}
\end{table}

\medskip\noindent\textbf{Compute details.}
\label{sec:compute}
Given the high resolution ($64800 = 180 \times 360$ sequence length from patch 4), we can only train most models with a local batch size of one. All our models are trained on NVIDIA A100 GPUs (40G memory capacity, unless specified). We find all models use most of the memory capacity, necessitating model parallelism at the higher scales.
For models up to $100M$, we use only data parallelism (16 GPUs, local batch size 1, global batch size 16). For models up to 304M, we use a $2\times1$ spatial parallelism (32 GPUs in total). For the large models, we observe training times of 4--6 hours. For example, the $204M$ model trains to \sci{6E19} FLOPs in about 6 hours on 32 GPUs.
For the largest model ($456.7M$ parameters) in neural scaling, we use $2\times2$ spatial parallel (4 GPUs) for a total of 64 GPUs. This model trains to \sci{6E19} FLOPs in about 3 hours.
For AR cooldowns with 4 steps, the memory complexity increases linearly and hence this model would then require 16 GPUs for spatial parallel and a total of 256 GPUs.
We estimate from the cluster summary usage that the neural scaling experiments (15 models, 57 different iteration counts for cooldowns in total) took about 4000 GPU hours, excluding the extrapolation runs, (several models were trained longer than they needed to be trained) with an additional O(2000) GPU hours for different hyperparameter experiments. The biggest compute expense was the $1.24B$ model.
For only the $1.24B$ parameter model, we use 80G A100s and train it on 64 GPUs with spatial parallel $4$. We do this since this training run is quite expensive. Our cluster contains 4-way NVLINK between GPUs and hence keeping spatial parallelism to less than 4 is ideal to avoid communication bottlenecks. This training run takes about 3.3 days to finish (to \sci{2.25E21} FLOPs; about 5000 GPU hours) on the 64 GPUs and hence is an expensive run. This is about a 40\% model FLOPs utilization (MFU) (compared to peak A100 Tensor Core FP16 FLOPs) which is reasonable for Transformer models.

\noindent \textbf{Hybrid spatial (domain) + data parallelism.}
The main ingredient in spatial parallelism is implementing a distributed roll operation.
Instead of shifting the windows directly, the original {\swin} implementation \citep{liu2021swintransformerhierarchicalvision} simply rolls (cyclically shifts) the image instead and applies an attention mask to limit the attention to each sub-window.
When the image is partitioned across GPUs, this roll operation must happen across the GPUs that shard the image.
\begin{algorithm}
\caption{Distributed Roll for Shifting Windows in {\swin}}
\label{alg:distributed_roll}
\begin{algorithmic}[1]
\REQUIRE Roll dimension $d$, local tensor $x$ (spatially sharded in $d$ across GPUs), shift size $s$, spatial parallel GPU group $\mathcal{G}$
\ENSURE Tensor $x$ rolled along dimension $d$

\STATE $(r, P) \gets \mathrm{rank\_and\_size}(\mathcal{G})$
\COMMENT{GPU rank and world size}

\IF{$P = 1$ \OR $s = 0$}
    \STATE \textbf{return} $\mathtt{roll}(x, s, d)$
    \COMMENT{purely local case}
\ENDIF

\STATE $\text{right} \gets (r + 1) \bmod P$
\STATE $\text{left}  \gets (r - 1) \bmod P$
\COMMENT{neighboring GPUs in the ring}

\IF{$s > 0$}
    \STATE $\text{send} \gets x[\dots, -s{:}, \dots]$
    \COMMENT{right boundary slice}
\ELSE
    \STATE $\text{send} \gets$  $x[\dots, {:}{-s}, \dots]$
    \COMMENT{left boundary slice}
\ENDIF

\STATE $\text{recv} \gets \mathtt{zeros\_like}(\text{send})$
\COMMENT{receive buffer}
\STATE {$\mathtt{distributed\_all\_to\_all\_single}(\text{recv}, \text{send}, \mathcal{G})$}
\COMMENT{$s>0$ sends right / receives left}\\
\COMMENT{$s<0$ vice versa}

\STATE $x \gets \mathtt{roll}(x, s, d)$
\COMMENT{local roll (boundary will be corrected)}

\IF{$s > 0$}
    \STATE $x \gets \mathtt{concat}(\text{recv},$  $x[\dots, s{:}, \dots]$$, d)$
\ELSE
    \STATE $x \gets \mathtt{concat}(x[\dots, {:}s, \dots], \text{recv}, d)$
\ENDIF

\STATE \textbf{return} $x$
\end{algorithmic}
\end{algorithm}
Hence, these spatial parallel GPUs must send and receive data from neighbors for an accurate roll. This boils down to a \emph{collective permute} operation and we implement it using {PyTorch}'s
{\fontfamily{cmtt}\bfseries torch.distributed.all\_to\_all\_single}
collective. Following frontier NLP research codes from
{\fontfamily{cmtt}\bfseries Megatron-LM} \citep{megatron-lm}, we implement the forward and backward pass of this distributed operation using PyTorch's custom {\fontfamily{cmtt}\bfseries Autograd} functions, where the backward pass is simply the conjugate forward operation: a reverse distributed roll.
We show the distributed roll algorithm in Alg. \ref{alg:distributed_roll}.
We note that
this is not a full replacement for distributed {\fontfamily{cmtt}\bfseries torch.roll},
as we assume we only need to send one slice to one
neighboring GPU and receive one slice from another neighboring GPU.
This is true for {\swin}, because we shift by a half of the window size, and windows are not sharded across GPUs in our setting (there is at least one window on every GPU).
For brevity, we also omit the {\fontfamily{cmtt}\bfseries permute} operations to bring the shift dimension to position 0 and back ({\fontfamily{cmtt}\bfseries all\_to\_all\_single} only acts on the first dimension).

Finally, spatial parallel introduces weights (and bias) tensors that are \emph{shared} across the spatial parallel GPU groups (in addition to the data parallel GPUs that, by design, share the model weights). Hence, the weight gradients ({\fontfamily{cmtt}\bfseries wgrads}) must be additionally reduced in this group (along with the data parallel group). Our simplistic {\swin} model design (with coordinate position embeddings) leads to all the weights (and bias) tensors shared across the spatial parallel group and hence we implement this as a joint {\fontfamily{cmtt}\bfseries AllReduce} operation across the combined data and spatial parallel GPU group using a communication hook in PyTorch's {\fontfamily{cmtt}\bfseries DistributedDataParallel}. We verify with Unit Tests that the distributed forward and backward passes (including weight gradients and adjoints) match the non-distributed version ensuring accuracy of our custom model parallelism.
Finally, we also use {\fontfamily{cmtt} \bfseries NVIDIA TransformerEngine} \citep{TransformerEngine} for easy access to fused operations and support for 4D parallelism with tensor parallelism.
While we do not use tensor parallelism in any run, we note that the code supports this parallelism as well for scaling up.
Here, an additional orthogonal array of GPUs may be employed to further shard the model weights (similar to NLP). {\fontfamily{cmtt}\bfseries TransformerEngine} provides a drop-in replacement for {\fontfamily{cmtt}\bfseries Linear} layers in our model to use tensor parallelism. We note that some {\fontfamily{cmtt}\bfseries bias} tensors may be shared across tensor and spatial parallel GPU groups with this additional parallelism and we take care of this through custom communication hooks in {\fontfamily{cmtt}\bfseries DistributedDataParallel}. Unit tests confirm that combined spatial, tensor, and data parallelism yield the correct distributed forward pass outputs and backward pass {\fontfamily{cmtt}\bfseries wgrads} and {\fontfamily{cmtt}\bfseries dgrads}.

\noindent \textbf{Efficient and stateful dataloaders.}
Our data is stored in HDF5 files, with one file per year of ERA5, to efficiently leverage the Lustre filesystem in our experiments for high-throughput data access. We use  {\fontfamily{cmtt}\bfseries NVIDIA DALI} \citep{nvidia_dali} dataloaders, which automatically overlap data loading with computation during the forward and backward passes and perform pre-processing operations, such as data normalization, directly on the GPU. This ensures that data I/O does not become a performance bottleneck.
Finally, we train our models to fixed iteration counts (to support periodic cooldowns under a predefined FLOP budget), which requires checkpoint--restart functionality to operate correctly mid-epoch. To enable this, we implement stateful dataloaders within DALI that track both the shuffled sample indices and the number of completed iterations, allowing training to resume seamlessly from arbitrary mid-epoch checkpoints.

\noindent \textbf{Code and FLOPs.}
\label{sec:flops}
We open-source our code at \url{https://github.com/ShashankSubramanian/neural-scaling-weather}.
For all our FLOP computations, due to the homogeneous design of the {\swin}, we use standard analytical attention and MLP FLOP estimates, taking into account the windowed attention, and verify with {\fontfamily{cmtt}\bfseries fvcore} \citep{fvcore} that our analytical values are close to the correct estimates. We include the patch, position, and reverse-patch embedding FLOPs as well in our estimation. The FLOP expressions can be found in {\fontfamily{cmtt}\bfseries utils/flops\_utils.py} in our code.

For comparison to GraphCast \cite{lam2023graphcast}, we use their open-sourced code base \url{https://github.com/google-deepmind/graphcast/blob/main/graphcast_demo.ipynb} to get a preliminary estimate on the FLOPs. We use the  \texttt{compiled.cost\_analysis()} from \texttt{JAX} to compute the FLOPs for a forward pass of the saved GraphCast checkpoint on the 0.25$^\circ$ ERA5 data. The saved checkpoint has $36.4M$ parameters (with latent dimension 512) and the FLOP estimate for a single forward pass (batch size 1) is \sci{2.97E13} FLOPs. We then multiply with the following training details reported by \citet{lam2023graphcast}: batch size 32 with $300,000$ iterations for pre-training. This results in \sci{2.85E20} FLOPs for the full forward pass. Assuming backward is $2\times$, the pre-training cost is \sci{8.55E20} FLOPs. In fine-tuning, GraphCast performs 2--12-step {\AR} over $11,000$ iterations ({\AR} increased by 1 every 1000 iterations) . Assuming that {\AR}-$N$ amplifies the FLOP count by $N$, this is equivalent to $ (2 + 3 + \cdots + 12) \times 1000 = 77 \times 1000 = 77,000$ 1-step forward passes in fine-tuning, or \sci{2.19E20} FLOPs. Roughly 20\% of the total compute was used for fine-tuning, with a total of around \sci{1E21} FLOPs. For the $204M$ {\swin}, we use approximately \sci{6E19} FLOPs in pre-training and with 8-step {\AR} in cooldowns (5\%), the total amounts to \sci{8.4E19} FLOPs.

\begin{figure}[t]
    \centering
    \includegraphics[width=1\linewidth]{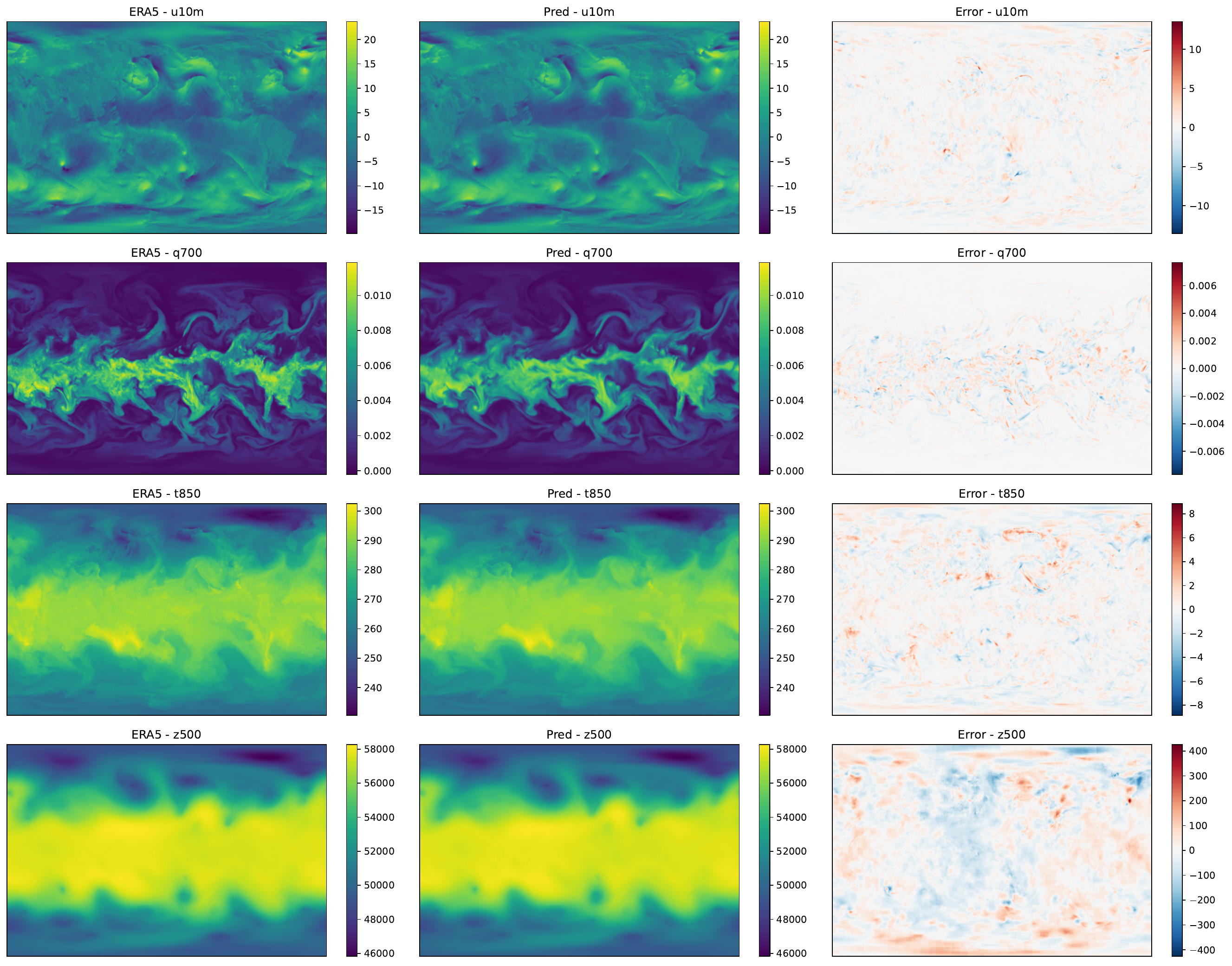}
    \caption{\textbf{Predictions from the 204M parameter model:} We show the ground truth ERA5, the prediction, and the error in physical units for four variables. The predicted forecasts also qualitatively match the ground truth.}
    \label{fig:viz}
\end{figure}

\subsection{Additional results and analyses}

\noindent \textbf{Prediction visualization.} We visualize the prediction from the compute-optimal model ($203.8M$) at the large budget \sci{6E19} in Fig. \ref{fig:viz}, along with the bias (error) between the predictions and the ground truth (ERA5).
The predictions are obtained on the testing year 2020, with initial conditions every 12 hours (for a total of 732 initial conditions).
Beyond the quantitative metrics defined earlier, we observe that the model produces realistic predictions of the physical fields, capturing the main spatial structures and patterns present in the reference data.

\noindent \textbf{Inference speed.} All the {\swin} models fit on a single GPU (A100 40G) during inference. Most models generate 10-day forecasts in less than a minute (including metric computations). For example, the $204M$ {\swin} takes 32 seconds, while the largest $1.24B$ {\swin} takes around 96 seconds to generate the same 10-day forecast.
These models preserve the emulator's substantial computational advantage over the NWP, achieving a speedup of more than 100$\times$ \citep{Alexe2024AIFS}.

\begin{figure}
    \centering
    \begin{subfigure}[b]{\textwidth}
        \centering
        \includegraphics[width=\textwidth]{figures/spectra_24h.pdf}
        \caption{PSD of different variables compared to PSD of the truth (ERA5) for different models at 24 hours lead time.}
        \label{fig:spec24h}
    \end{subfigure}
    \hfill
    \begin{subfigure}[b]{\textwidth}
        \centering
        \includegraphics[width=\textwidth]{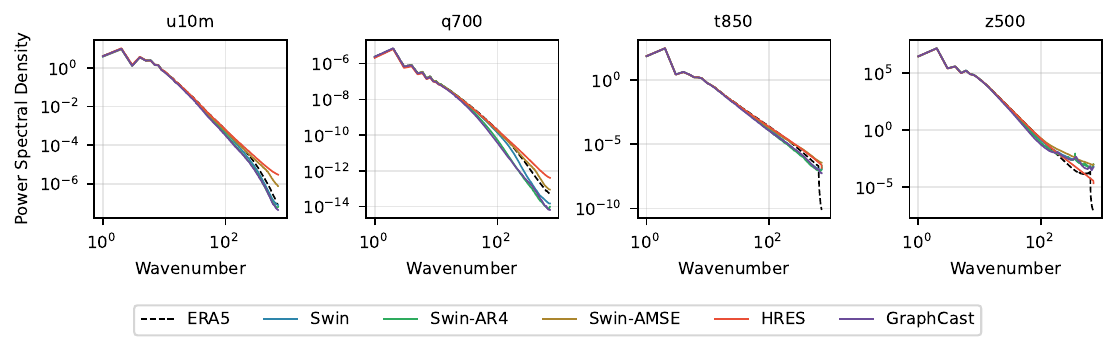}
        \caption{PSD of different variables compared to PSD of the truth (ERA5) for different models at 120 hours lead time.}
        \label{fig:spe120h}
    \end{subfigure}
    \caption{\textbf{PSD of predictions at 1 day and 5 days: }
    While {\AR} produces blurrier forecasts (though with lower RMSE), {\AMSE} helps retain power at high wavenumbers. At longer lead times, such as 120 hours, {\swin}-{\AMSE} continues to show significant power at these small scales. This allows practitioners to define an effective resolution based on this noise-level as outlined by \citet{subich2025fixingdoublepenaltydatadriven}.}
    \label{fig:spectra-additional}
\end{figure}

\noindent \textbf{Cooldowns and Alignment.} In Fig. \ref{fig:spectra-additional}, we show the PSD of different variables for two lead times (1 day and 5 days). The {\swin} model cooled down with {\AR}4 loses a significant amount of energy at small scales, resulting in overly dissipated forecasts and an effective resolution much coarser than what the $0.25^\circ$ grid can support. In contrast, the {\swin} model cooled down with {\AMSE} matches the PSD of the ground truth almost exactly at short lead times. At longer lead times, it exhibits higher energy than the ground truth at the smallest scales suggesting a noise-driven effective resolution that exceeds that of the dissipated {\AR}4 forecasts, as noted by \citet{subich2025fixingdoublepenaltydatadriven}. Overall, the two cooling strategies guide the model toward different objectives: one prioritizes lower overall RMSE (and skill) over long lead times, akin to ensemble forecasting, while the other emphasizes high realism, preserving small-scale features in the forecast.

\begin{figure}
    \centering
    \includegraphics[width=1\linewidth]{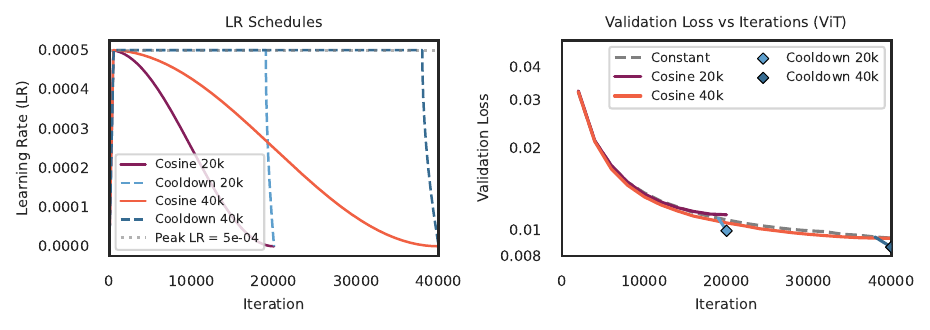}
    \caption{\textbf{Cosine vs cooldowns for a Vision Transformer (ViT):} We train a simple ViT (patch size 8, embedding 768, depth 12, LR \sci{5E-4}, learnable position embeddings, batch size 16) using both cosine schedules (to 20,000 and 40,000 iterations) and a constant + cooldown schedule. The continuous training shows that the cooldowns are superior to the cosine schedule for this ViT as well. This preliminary evidence points to the possibility that the cooldown methodology is  agnostic to the architecture.}
    \label{fig:vit}
\end{figure}

\begin{figure}
    \centering
    \includegraphics[width=1\linewidth]{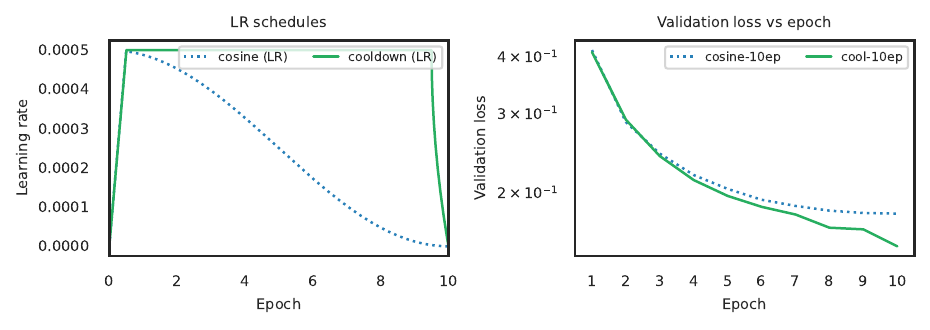}
    \caption{\textbf{Cosine vs cooldowns for FourCastNet3:} We train a small FourCastNet3 model using the deterministic training framework from the \href{https://github.com/NVIDIA/makani}{Makani} repository. Each pressure, surface, and auxiliary variable is embedded with a dimension of 12; and we use the default 360 x 720 internal resolution (scale factor 2). Since the repository supports training and logging at the epoch level, we track validation loss after each epoch. To keep training manageable, we use 16,000 samples per epoch with a batch size of 16. We compare two learning rate schedules: cosine decay and a constant+cooldown schedule. In this preliminary experiment, the constant+cooldown schedule proves beneficial  for Neural Operator-based models as well such as FourCastNet3 (Spherical Fourier Neural Operator). These initial results suggest that the cooldown methodology may be largely architecture-agnostic.}
    \label{fig:fcn}
\end{figure}

\noindent \textbf{Cooldowns vs. cosine. } With some preliminary experiments, we show that the performance improvement of cooldowns w.r.t cosine can be true for other architectures as well. In Fig. \ref{fig:vit}, we demonstrate this for a simple Vision Transformer and in Fig. \ref{fig:fcn}, we show this for a SOTA model, FourCastNetv3.

\noindent \textbf{Channel, area weighting, and AMSE. }Other domain-specific modifications during cooldown can also be done. For example, models commonly weight the loss to emphasize certain variables or regions through channel and area weighting.
We employ the channel weighting in GraphCast \cite{lam2023graphcast} which down-weights the loss as the pressure level decreases and also introduced surface variable weights and the standard cosine latitude area weighting (similar to the RMSE) for area weights.
We train a {\swin} model with weighted losses in pre-training and compare with MSE pre-training and cooling down with the weighted losses.
We report different ablations in Tab. \ref{tab:chwt}.
We observe that the cooldown can be repurposed with channel or area weighting with similar results as pre-training from scratch, demonstrating that the neural scaling can be disentangled from channel weighting as well.
We also see that channel weighting generally reduces the loss and area weighting increases it.
Similarly, AMSE-only cooldowns yield similar losses as training AMSE from scratch; both variants preserve high-resolution features in the PSD.
From-scratch AMSE training sometimes produced noisy pixels which we were able to mitigate by adding a residual connection from the input to the output; we did not observe the instability in the AMSE cooled down models.
This supports using AMSE for adaptation, consistent with \citet{subich2025fixingdoublepenaltydatadriven}, while reusing MSE-pre-trained checkpoints.

\begin{table}
\caption{\textbf{Pre-training versus cooldown objectives.} All runs use the 115M model trained for 40,435 iterations (closest compute-optimal for \sci{3E19} FLOPs). CHWT and AWT denote channel and area weighting of MSE. AMSE loss values are not directly comparable to the MSE-based losses.}
\label{tab:chwt}
\centering
\small
\begin{tabular}{llr}
\toprule
Pre-training objective & Cooldown objective & Validation loss \\
\midrule
MSE & MSE & 0.0068 \\
MSE & MSE (CHWT) & 0.0062 \\
MSE (CHWT) & MSE (CHWT) & 0.0061 \\
MSE & MSE (CHWT + AWT) & 0.0075 \\
MSE (CHWT + AWT) & MSE (CHWT + AWT) & 0.0072 \\
\midrule
AMSE & AMSE & 0.109 \\
MSE & AMSE & 0.119 \\
\bottomrule
\end{tabular}
\end{table}

\begin{figure}
    \centering
    \includegraphics[width=1\linewidth]{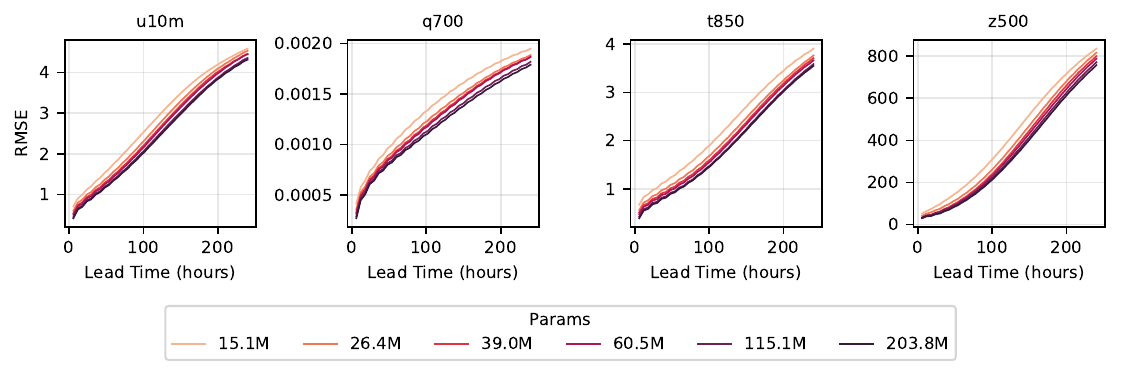}
    \caption{\textbf{Cooldown with {\AR}4 is valid for different model sizes.} {\AR}4 gains consistently improve with model size (and compute), similar to Fig. \ref{fig:rmse-flops} that shows the same improvements in the {\AR}1.}
    \label{fig:cooldown-ar4-valid}
\end{figure}

\begin{figure}
    \centering
    \includegraphics[width=1\linewidth]{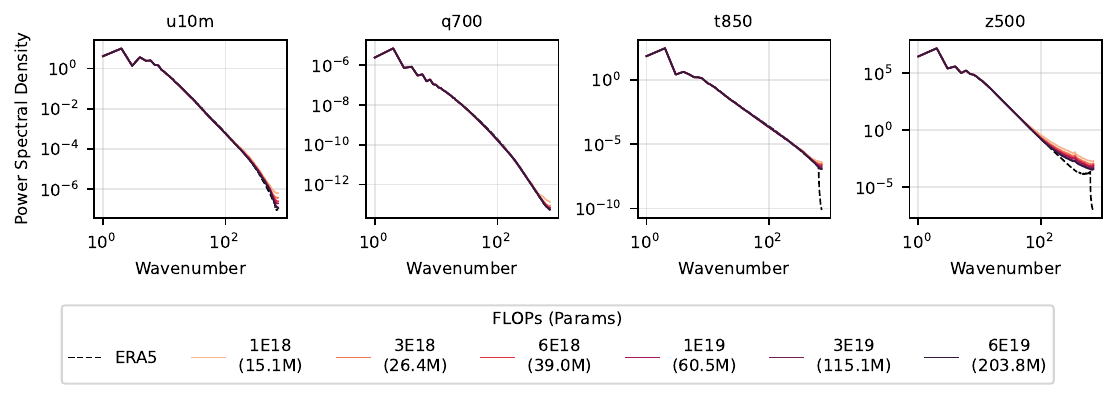}
    \caption{\textbf{Cooldown with {\AMSE} is valid for different model sizes.} {\AMSE} allows all models to match the ERA5 PSD more closely. PSD is computed at 24h lead time.}
    \label{fig:cooldown-amse-valid}
\end{figure}

\begin{figure*}
    \centering
    \includegraphics[width=1\linewidth]{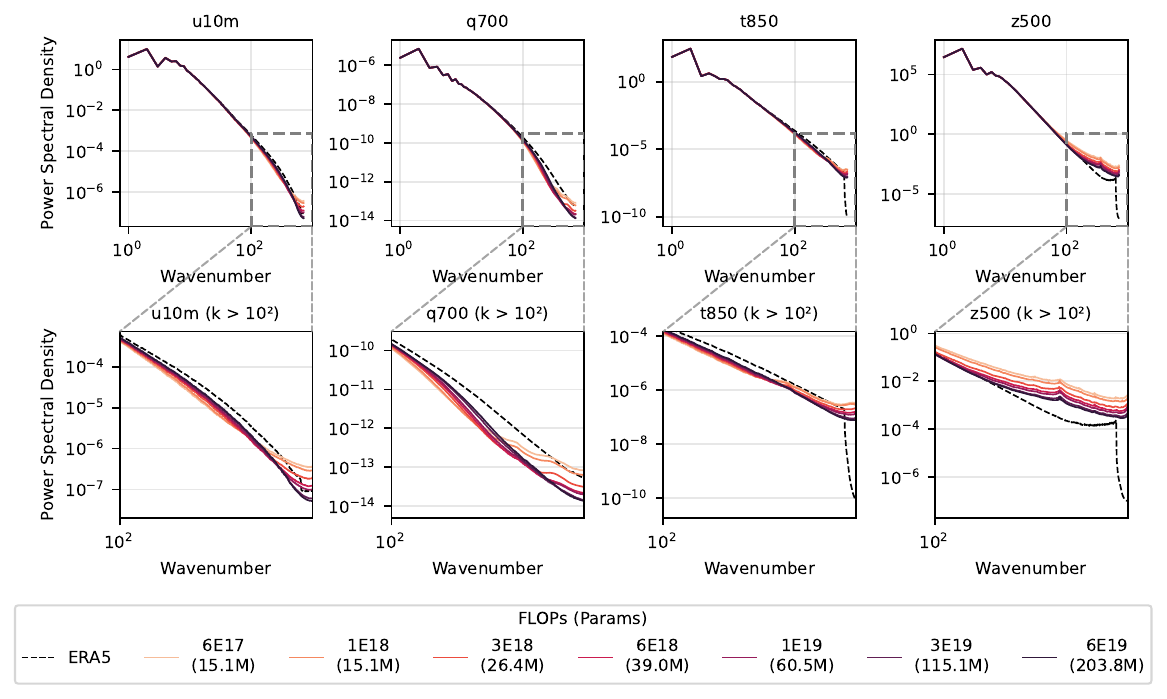}
    \caption{\textbf{PSD performance at lead time 24 hours of the closest compute-optimal {\swin} models at each compute budget:} As in Fig. \ref{fig:rmse-flops} that shows consistent improvement of rollout RMSE with increased compute, the PSD also exhibits improvements, though they are smaller in magnitude and progress more gradually. Here, the PSD is at 24-hour lead time (averaged over the test year). In the bottom row, we zoom into the high wavenumbers and can observe that the bigger (more compute) models show sharper features and resemble ERA5 closer.}
    \label{fig:spectra-flops}
\end{figure*}

\begin{figure*}
    \centering
    \includegraphics[width=1\linewidth]{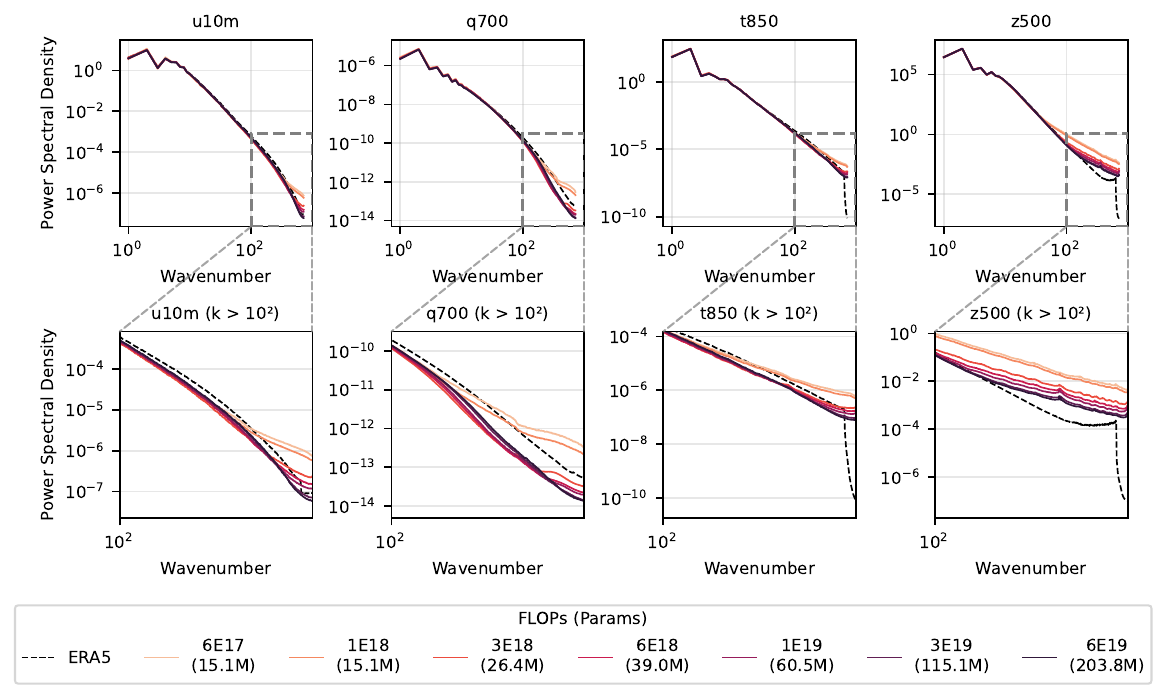}
    \caption{\textbf{PSD performance at lead time 120 hours of the closest compute-optimal {\swin} models at each compute budget:} The PSD is evaluated at a 120-hour lead time and averaged over the test year. All models accurately capture the low wavenumbers, while higher wavenumbers show gradual improvement as compute increases. The bottom row zooms in on the high-wavenumber range, where larger models exhibit spectra that more closely match ERA5, whereas smaller models appear more blurred and display some artificial high-wavenumber features. These features are more dominant at this longer lead time compared to 24 hours in Fig. \ref{fig:spectra-flops}.}
    \label{fig:spectra-flops-120h}
\end{figure*}

\noindent \textbf{Downstream performance and compute.}
In Fig. \ref{fig:rmse-flops}, we present the RMSE over the time horizon using the closest compute-optimal model at each budget.
As the compute increases, accuracy improves for both one-step predictions (as shown in Fig. \ref{fig:scaling-model}) and over longer time horizons. At the highest budgets, however, we start to see a saturation.
Fig. \ref{fig:cooldown-ar4-valid} and Fig. \ref{fig:cooldown-amse-valid} show that repurposing cooldowns with either {\AR} or {\AMSE} is valid across different model sizes (and compute). This demonstrates the generality of the approach, allowing models to be aligned with different downstream objectives during the cooldown period.
In Fig. \ref{fig:spectra-flops} and Fig. \ref{fig:spectra-flops-120h}, we show the PSD for the closest compute-optimal model at each compute budget for a 24-hour and 120-hour lead time. While all models faithfully capture the PSD at the lower wavenumbers, the higher compute models show better PSD at the higher wavenumbers as well, better capturing the sharper features.
However, improvements with increasing compute are gradual and show signs of saturation. Additionally, lower-compute models exhibit artificial power at high wavenumbers, which becomes more pronounced at longer lead times (120 hours). In contrast, higher-compute models maintain more consistent performance across lead times.


\begin{figure*}
    \centering
    \includegraphics[width=1\linewidth]{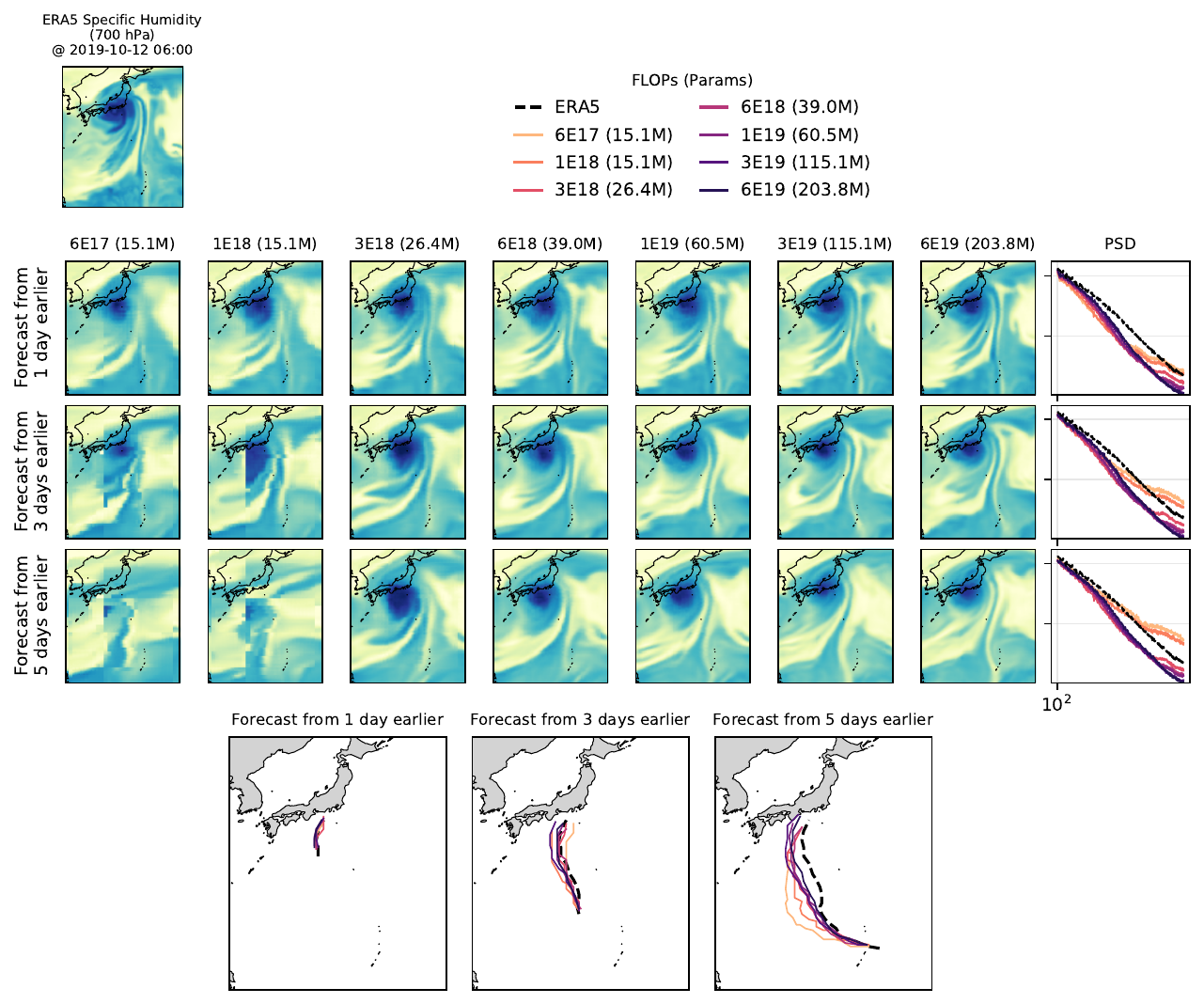}
    \caption{\textbf{Tropical Cyclone (TC) visualization and tracks before Typhoon Hagibis made landfall in Japan}: We reproduce the TC analysis of \citet{price2023gencast}. We show the ERA5 ground truth for specific humidity at 700hPa on October 12, 2019, 06 UTC in the top left. The following three rows show the {\swin} predictions at the same time from all the closest compute-optimal models (between \sci{6E17} and \sci{6E19} FLOPs), initialized 1, 3, and 5 days earlier, respectively. The last column shows the PSD for the predictions (zoomed into the high wavenumbers). Forecasts initialized 1 day earlier show that all models capture the TC structure, although the larger models produce systematically sharper features, which is also reflected in the PSD. For initializations 3 and 5 days earlier, the smaller models fail to capture the TC structure, exhibiting patching artifacts and generally inconsistent blending of physical features. This behavior is likewise evident in the PSD as artificial high-wavenumber energy. In contrast, the larger models consistently capture the TC across all lead times. The last row shows the TC track from the different initializations. Although a deterministic model is insufficient for accurately assessing track performance, some variability among the different models is seen.}
    \label{fig:tc}
\end{figure*}

These spectral improvements are more visible in a specific physical case study in Fig. \ref{fig:tc}.
We reproduce the study that analyzes the Tropical Cyclone (TC) patterns of Typhoon Hagibis by \citet{price2023gencast}. We use the same TC tools as the study---TempestExtremes \cite{ullrich2021tempestextremes}, in order to visualize TC tracks. Here, we track the TC structure across the compute-optimal models initialized at 1, 3, and 5 days from the TC landfall time on October 12, 2019, 06 UTC. We visualize the specific humidity at 700hPa for all the forecasts as well as the PSD (compared to ERA5). For short forecasts (1 day), all models are faithful with increased compute resulting in sharper predictions (also quantified in the PSD). With longer forecasts (3 and 5 days earlier), the smaller models exhibit artificial features (patch artifacts, discontinuities in the prediction) that are also seen in the spectral power at the high wavenumbers. With more compute, the models begin to produce realistic TCs at these lead times as well.
Although different hyperparameters might improve the smaller models, the artifacts are more consistent with undertraining or limited capacity. The two runs with window artifacts in Fig.~\ref{fig:tc} were cooled down before 10,000 iterations, while validation loss was still decreasing significantly; the same models trained longer do not show these artifacts.
We also visualize the TC tracks. While all models follow the general track of ERA5, the smaller models show more variability at longer lead times due to artifacts developing.
We note that, due to the deterministic modeling framework, these results should be interpreted as illustrative. Accurately tracking such phenomena, particularly at longer lead times, requires probabilistic (ensemble) forecasting, as the chaotic nature of atmospheric processes naturally introduces more spread (variance) in the predictions, particularly at longer lead times.

\begin{table}
\caption{\textbf{Patch size, embedding, depth ablations}. We vary patch size, embedding, and depth to keep models approximately on an IsoFLOP budget. All models are trained and cooled down to 20,000 iterations. Results show that (i) patch size 4 achieves the lowest loss, but the variance amongst patch sizes is not too high, and (ii) embedding and depth are largely interchangeable unless taken to extremes (very shallow or very deep), which increases loss.}
\centering
\small
\begin{tabular}{|c|c|c|c|c|c|c|c|c|}
\hline
Patch Size & Embedding & Depth & Window Size & TFLOPs/Iter & \#Params & LR &  Val. Loss \\
\hline
2 & 512  & 10 & (18, 36) & 20.0 & 31.79  & \sci{1.0E-3}  & 0.0085  \\
2 & 384  & 18 & (18, 36) & 21.4 & 32.11  & \sci{1.0E-3}  & 0.0083 \\
2 & 256  & 40 & (18, 36) & 23.5 & 31.65  & \sci{1.0E-3}  & 0.0086  \\
2 & 768  & 5  & (18, 36) & 21.2 & 35.86  & \sci{1.0E-3}  & 0.0091 \\
\hline
4 & 1024 & 12 & (9, 18)  & 20.4 & 153.49 & \sci{5.0E-4}    & 0.0075 \\
4 & 768  & 22 & (9, 18)  & 21.2 & 157.62 & \sci{5.0E-4}    &  0.0074 \\
4 & 1536 & 5  & (9, 18)  & 19.2 & 145.25 & \sci{5.0E-4}    &  0.0086\\
4 & 512  & 48 & (9, 18)  & 20.8 & 152.31 & \sci{5.0E-4}    & 0.0077 \\
\hline
8 & 2048 & 12 & (5, 10)  & 20.3 & 623.62 & \sci{2.5E-4}  &  0.0084\\
8 & 1536 & 22 & (5, 10)  & 20.8 & 637.64 & \sci{2.5E-4}  &  0.0081\\
8 & 1024 & 44 & (5, 10)  & 20.1 & 613.94 & \sci{2.5E-4}  &  0.0085 \\
8 & 768  & 88 & (5, 10)  & 20.6 & 630.45 & \sci{2.5E-4}  &  0.0089    \\
\hline
\end{tabular}
\label{tab:patch}
\end{table}

\noindent \textbf{Patch size and model aspect ratio ablation. }
To determine the appropriate patch size for the neural scaling experiments, we conduct an ablation study in Tab. \ref{tab:patch} in which we vary patch size, embedding, and depth to keep models approximately on an IsoFLOP budget.
Smaller patches use fewer parameters due to increased compute from longer sequences, while larger patches use more parameters; embedding and depth are also balanced to consume similar compute where we can train wider or deeper models. The parameter count includes the patch embedders and head. We also adjust the window size to maintain constant physical window area, so larger patches use proportionally smaller windows---this makes the IsoFLOPs slightly approximate as seen in the TFLOP per iteration values. All models are trained and cooled down to 20,000 iterations. Our results show that (i) patch size 4 achieves the lowest loss, but the variance amongst patch sizes is not too high, and (ii) embedding and depth are largely interchangeable unless taken to extremes (very shallow or very deep), which increases loss. We find a slight preference for deeper models, differing from the preference for wider models in the smaller compute range studied by \citet{yu2026scaling}. Our minimalist architecture and higher compute budgets are plausible explanations: narrow embeddings may impose a stronger information bottleneck at smaller scales, whereas additional depth may be more useful once that bottleneck is reduced. We do not isolate these effects here. Hence, we select patch size 4 for our scaling experiments. We note that resolution scaling remains an important future direction where this ablation is repeated across data, model sizes, and compute.

\label{sec:sens}
\noindent \textbf{Neural scaling sensitivity. }
For the neural scaling law analysis, we fit a quadratic curve to the loss for each IsoFLOP setting as a function of log-parameters or log-samples (similar to standard neural scaling analyses \cite{hoffmann2022trainingcomputeoptimallargelanguage}):
\[
L = p(\log N)^2 + q(\log N) + r,
\]
using a second-order polynomial fit (e.g., \texttt{polyfit} with degree 2). We include all loss values in the fit (no outlier removal) and use the resulting expression to estimate the local minimum. The quadratic-fit design matrices have condition numbers ranging from $13$ to $798$, indicating no severe numerical ill-conditioning. The coefficient of determination ($R^2$) ranges from $0.951$ to $0.998$ across the first eight IsoFLOP settings; the two highest budgets, \sci{3E19} and \sci{6E19}, yield $R^2=0.883$ and $0.836$, respectively. Although these values still indicate strong fits, the decrease suggests increasingly similar behavior among larger models and longer training runs. This trend is also reflected in the fitted coefficients: the quadratic term $p$ decreases with increasing compute, corresponding to flatter parabolas; a phenomenon also observed in LLMs~\cite{grattafiori2024llama3herdmodels}. The power-law fits to the compute-optimal points yield $R^2=0.9974$ for model size and $R^2=0.9668$ for loss, evaluated in log--log space.

\begin{figure}[t]
    \centering
    \includegraphics[width=\linewidth]{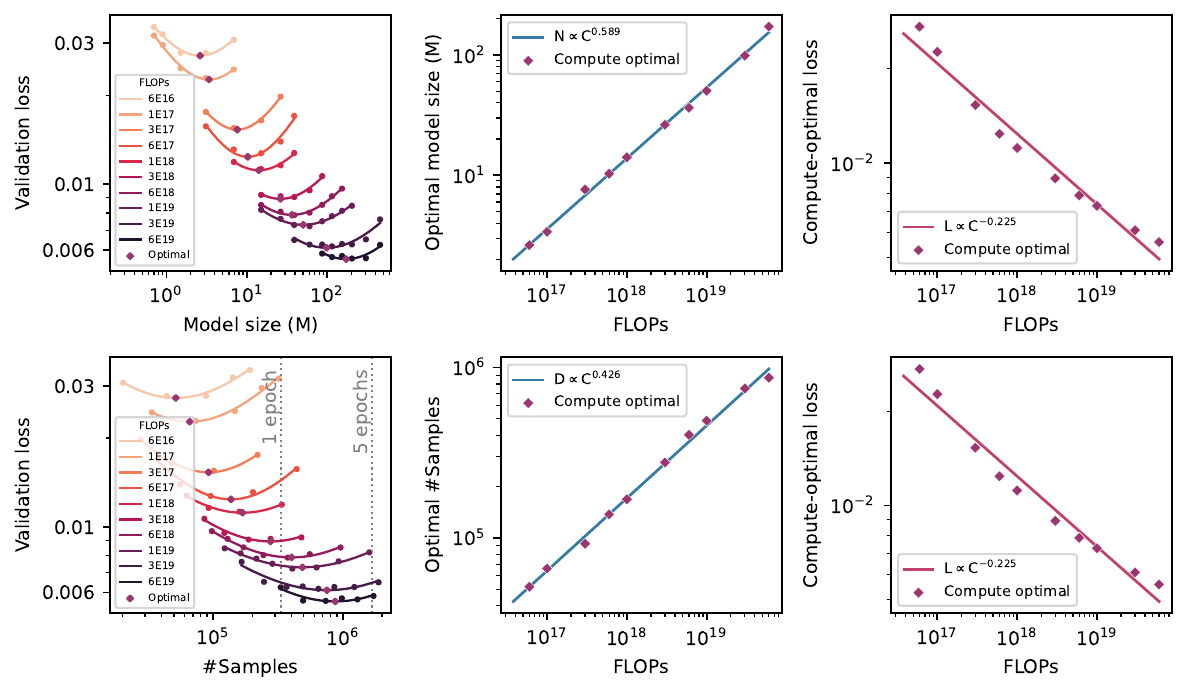}
    \caption{\textbf{Neural scaling with channel weighting loss function.}  We cool down all models using the GraphCast channel weights in the loss function that includes both weights based on pressure level as well as surface variable weighting.
    We find that the overall exponents  do not change too much (see legends for each scaling law). The compute-optimal models are all approximately the same. Loss values have slightly improved consistently due to the channel weights (also visible in the y-range).}
    \label{fig:scaling-model-new-chwt}
\end{figure}

\begin{figure}
    \centering
    \includegraphics[width=0.75\linewidth]{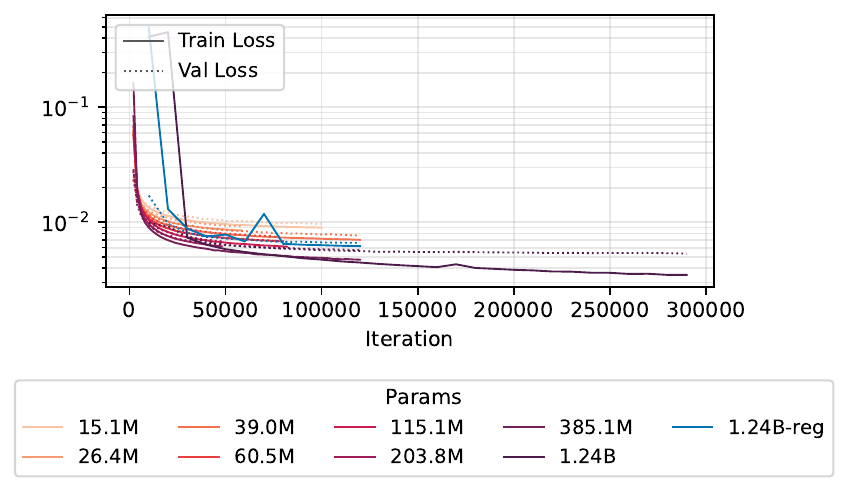}
    \caption{\textbf{Training and validation loss as a function of iterations for different models:} We show the training curves for the closest compute-optimal models, including the 385M intermediate-budget run and the 1.24B extrapolation run. Solid and dotted lines indicate training and validation loss, respectively. The extrapolation runs develop train--validation gaps as validation improvements diminish. The blue curve (1.24B-reg) uses stronger regularization (dropout 0.1 and weight decay 0.1), reducing the train--validation gap but plateauing at a higher validation loss; training was stopped after approximately 120K iterations.}
    \label{fig:loss-iter}
\end{figure}

\noindent \textbf{Neural scaling across multiple epochs. }\label{app:data-reuse}
We note that multiple epochs are necessary for a robust scaling analysis on this dataset. From Fig.~\ref{fig:scaling-data}, we observe that for many IsoFLOP settings, the characteristic U-shape of the parabola becomes visible only when including data points beyond one epoch. Restricting the analysis to less than one epoch limits the number of IsoFLOP regimes where a clear U-shaped curve can be identified. In such cases, only the left portion of the parabola is observed, requiring \textit{extrapolation} to estimate the optimal point. This significantly reduces the stability and reliability of the analysis. Therefore, we emphasize the use of multiple epochs to ensure that the parabolic trend is captured through interpolation across observed data points.
Another challenge arises when reducing compute budgets below \sci{6E17} in order to obtain more points within the sub-epoch regime. This necessitates training smaller ($<1$M parameter) models. However, such small models require many iterations to reach a given compute budget, often exceeding one epoch and thereby forcing the use of even smaller models. To capture both sides of the parabola under these constraints, larger models must be trained for very short durations (on the order of $\sim$1000 iterations), which approaches the warmup period (500 iterations). As a result, the models operate in an \emph{undertrained} or \emph{small-model} regime, where training dynamics and architectural choices (e.g., very shallow depths) have a disproportionate impact. This makes it challenging to reliably probe smaller compute budgets.
For these reasons, we focus on a compute range spanning three orders of magnitude, from \sci{6E16} to \sci{6E19}, where stable and consistent scaling trends can be observed. With larger and more diverse datasets (e.g., spanning multiple Earth science domains or cross-domain corpora), it may be possible to extend this range further, similar to LLMs.

\noindent\textbf{Variable-specific scaling. }\label{app:variable-scaling} All 71 channels exhibit clear compute scaling, with strong loss--compute fits (median $R^2=0.969$, minimum $0.924$). For $L\propto C^\gamma$, wind and temperature have exponents ($\gamma\approx -0.24$ to $-0.25$) relatively close to the aggregate loss exponent of $-0.217$. Pressure and geopotential scale fastest ($-0.39$ to $-0.43$), while moisture generally scales more slowly ($-0.19$ to $-0.21$). The smoother spatial structure of geopotential and the fine-scale structure of moisture are plausible explanations, but we do not establish their causes here. Vertical variation in scaling rates is modest for wind and geopotential but more pronounced for temperature and humidity; across model sizes, the distribution of error across pressure levels remains qualitatively similar. These exploratory findings suggest differences in variable-dependent predictability and could motivate channel weighting during cooldown based on scaling rates. Establishing the roles of spatial or temporal resolution, physical processes, and observational uncertainty requires a dedicated analysis, which we leave to future work.

\noindent \textbf{Neural scaling sensitivity to channel weighting. }
We select channel weighting as a popular training strategy to reduce the loss (as observed in Tab. \ref{tab:chwt}). We demonstrate that neural scaling can be repeated with channel weighting in the cooldown period. We repeat the full IsoFLOP curves with channel weighting in Fig. \ref{fig:scaling-model-new-chwt} where each model is cooled down with channel weights. We observe that while the validation loss has improved uniformly across all the models (due to channel weighting), the compute-optimal trends are very similar. The exponents for optimal model size vs. compute have changed from $0.584$ to $0.589$ with slightly different scaling constant, and for loss vs. compute has changed from $-0.217$ to $-0.225$. Hence, the exponents can be a little sensitive to the choice of loss, but the larger point remains: the models still show compute-optimal scaling behavior over three orders of magnitude of compute.

\noindent \textbf{Extrapolation analysis. } The $385M$ and $1.24B$ extrapolation runs fall short of their predicted losses (Fig.~\ref{fig:scaling-model}) after approximately 6 and 14 epochs, respectively, with train--validation gaps consistent with overfitting in this data-reuse regime.
We show this in Fig. \ref{fig:loss-iter} where for all the models in the scaling regime (\sci{6E17} to \sci{6E19}) the training and validation loss track closely.
However, for the models extrapolated to \sci{3E20} FLOPs and \sci{2.25E21} FLOPs, while the training loss reaches close to the extrapolated loss values, the validation loss plateaus at around $0.005$, pointing to significant overfitting.
We also trained a 1.24B model with stronger regularization (dropout 0.1 and weight decay 0.1). This reduced the train--validation gap, but validation loss plateaued near 0.006, slightly above the original run, so we stopped training after approximately 120K iterations. Stronger regularization therefore reduced overfitting without recovering the extrapolated scaling trend. Together with the intermediate-budget result, this supports finite data as the most likely limitation in our experiments, although gains from further hyperparameter tuning cannot be ruled out.

\begin{figure}
    \centering
    \includegraphics[width=1\linewidth]{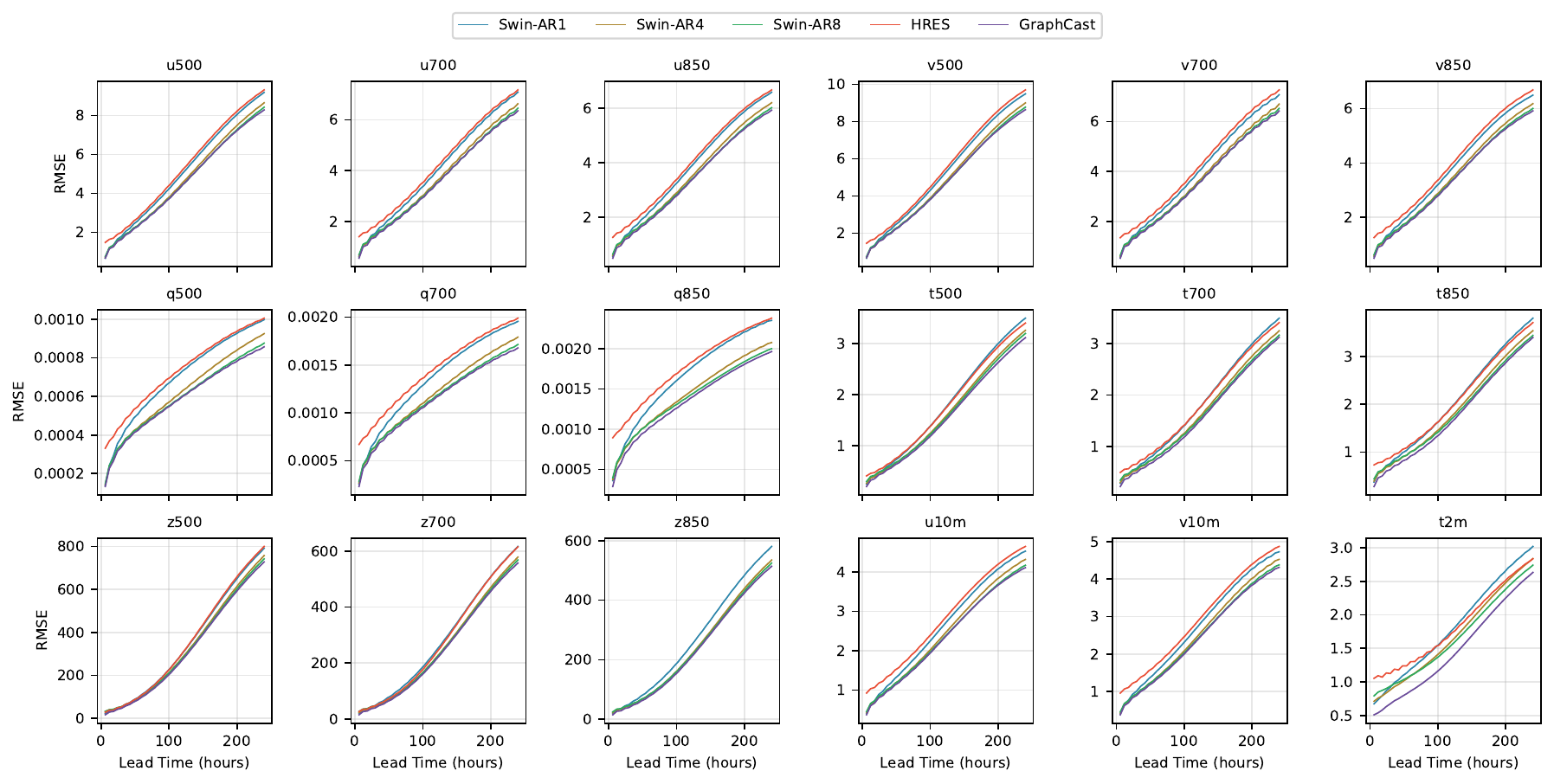}
    \caption{\textbf{RMSE vs time for a larger subset of variables.} Reproducing Fig. \ref{fig:rmse-opt} with more variables. Generally, we observe that the single step 204M {\swin} (compute-optimal \sci{6E19} FLOPs) outperforms HRES. With 4-step {\AR}, the {\swin} starts to approach GraphCast. With 8-step {\AR} cooldowns instead, the performance is similar to GraphCast. We note that (i) there was no data for z850 on HRES from gcsfs and (ii) the {\swin}'s performance on only 2m temperature is sub-optimal, although it shows similar behavior of approaching GraphCast performance with more {\AR}. This motivates custom channel weighting (different from the tuned weights for GraphCast) for specific variables.}
    \label{fig:rmse-all}
\end{figure}

\begin{figure}
    \centering
    \includegraphics[width=1\linewidth]{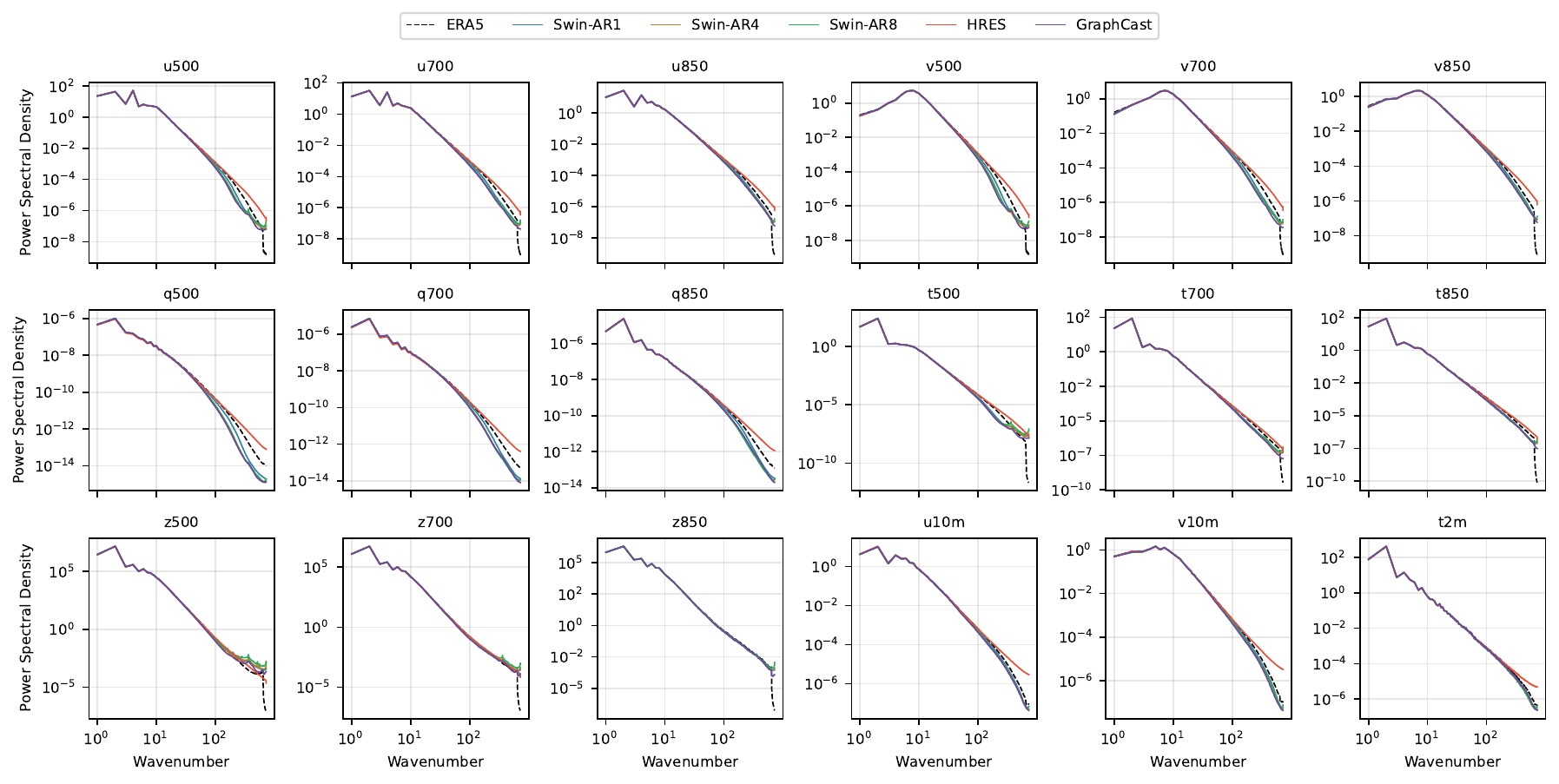}
    \caption{\textbf{PSD for a larger subset of variables at 24-hour lead time.}  As with the main paper figures, we observe that with more {\AR} the {\swin} also achieves the lower RMSEs observed in the above Fig. \ref{fig:rmse-all} through a smoother forecast. This is more visible at longer lead times (see next figure).}
    \label{fig:spectra-all-24h}
\end{figure}

\begin{figure}
    \centering
    \includegraphics[width=1\linewidth]{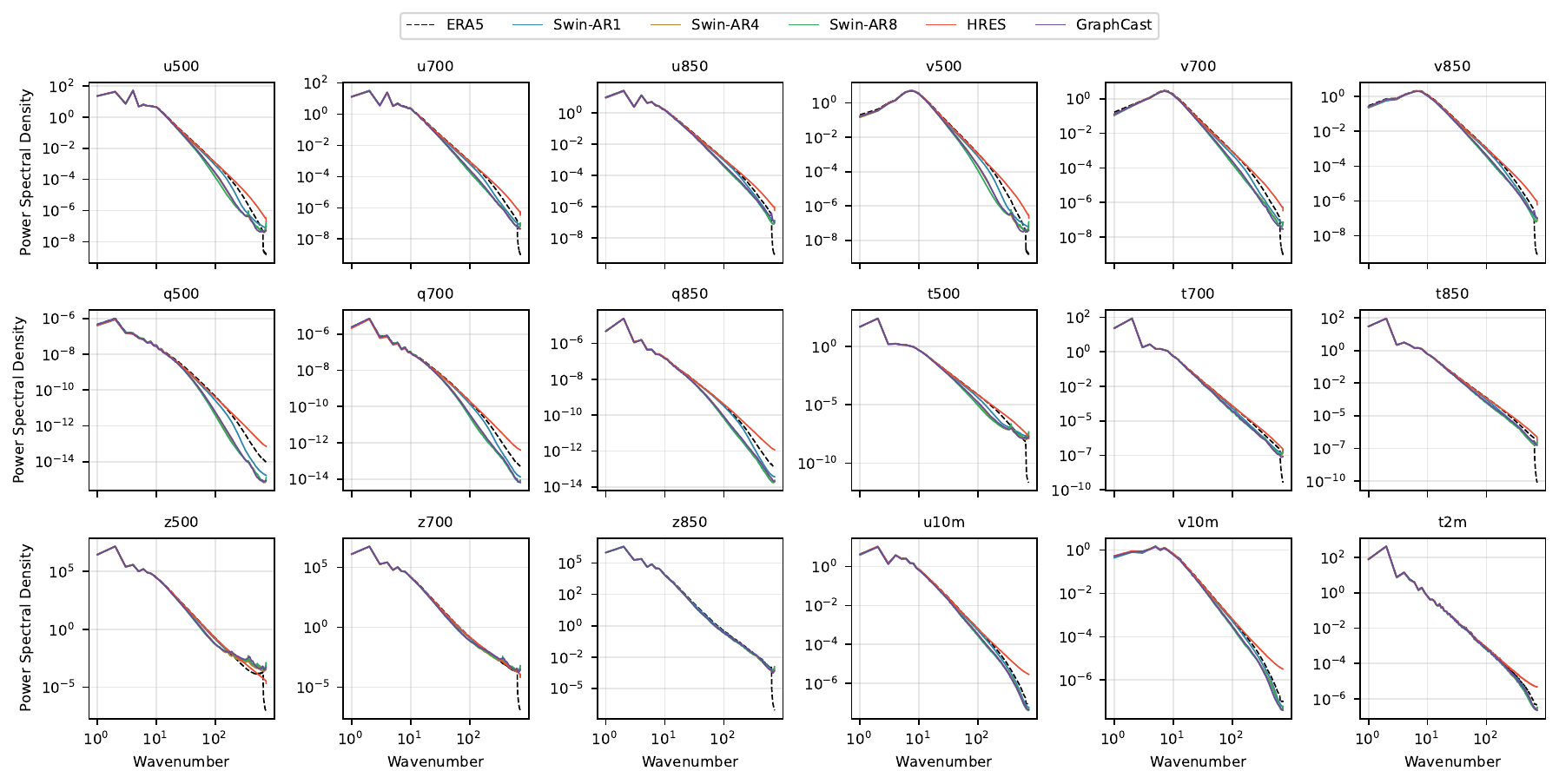}
    \caption{\textbf{PSD for a larger subset of variables at 120 hour lead time.}  At 120h, we can clearly see how both GraphCast and {\swin}-{\AR}8 over smooth to achieve the low forecast errors, thereby emulating an ensemble forecast. This is particularly visible in the specific humidity variables such as q700 (second row, second column). The one-step {\swin} shows lesser smoothing.}
    \label{fig:spectra-all-120h}
\end{figure}

\begin{figure}
    \centering
    \includegraphics[width=1\linewidth]{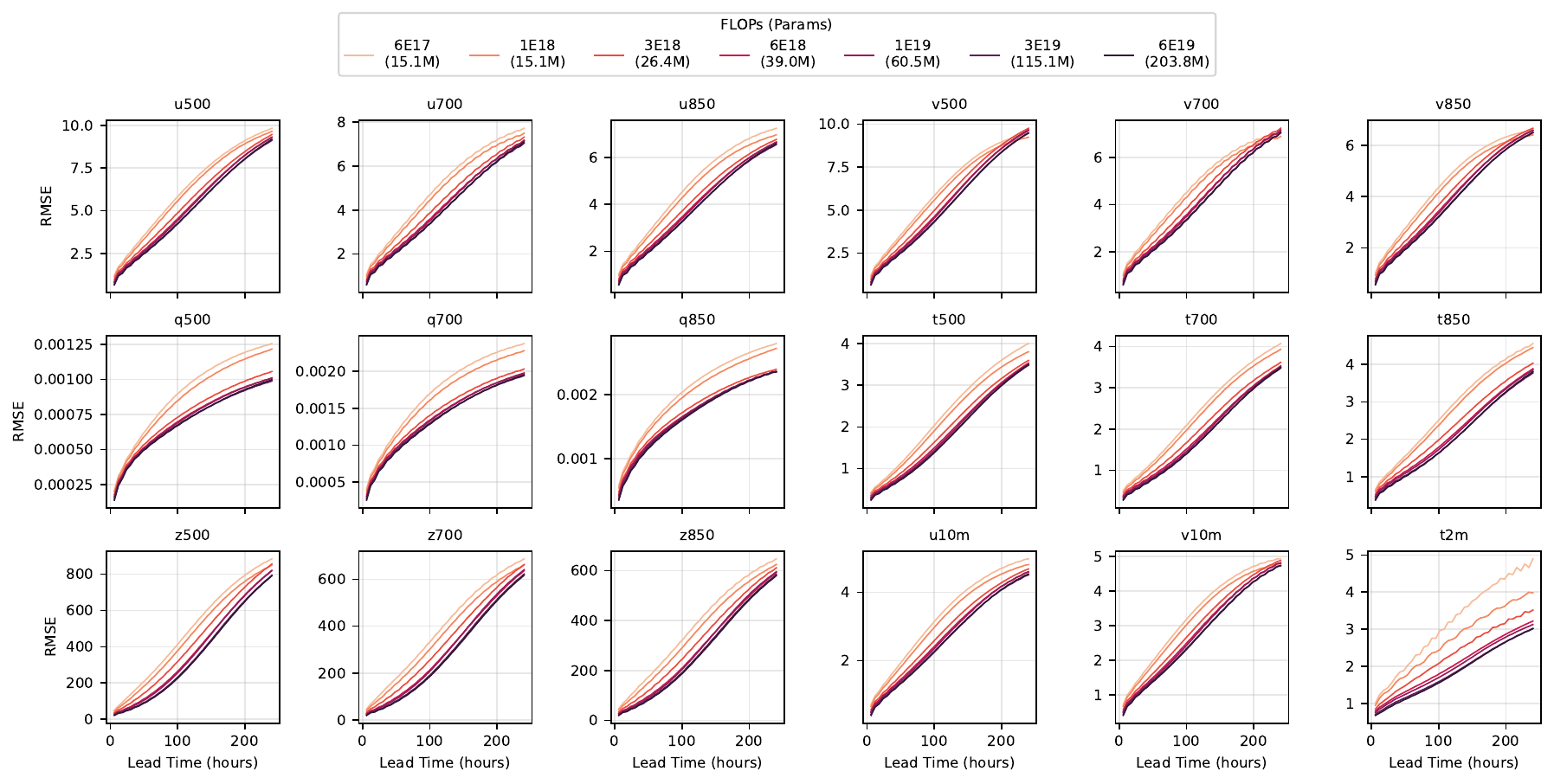}
    \caption{\textbf{RMSE performance of the closest compute-optimal {\swin} models at each compute budget for a larger subset of variables.} We show Fig. \ref{fig:rmse-flops} for more variables.}
    \label{fig:rmse-flops-all}
\end{figure}

\begin{figure}
    \centering
    \includegraphics[width=1\linewidth]{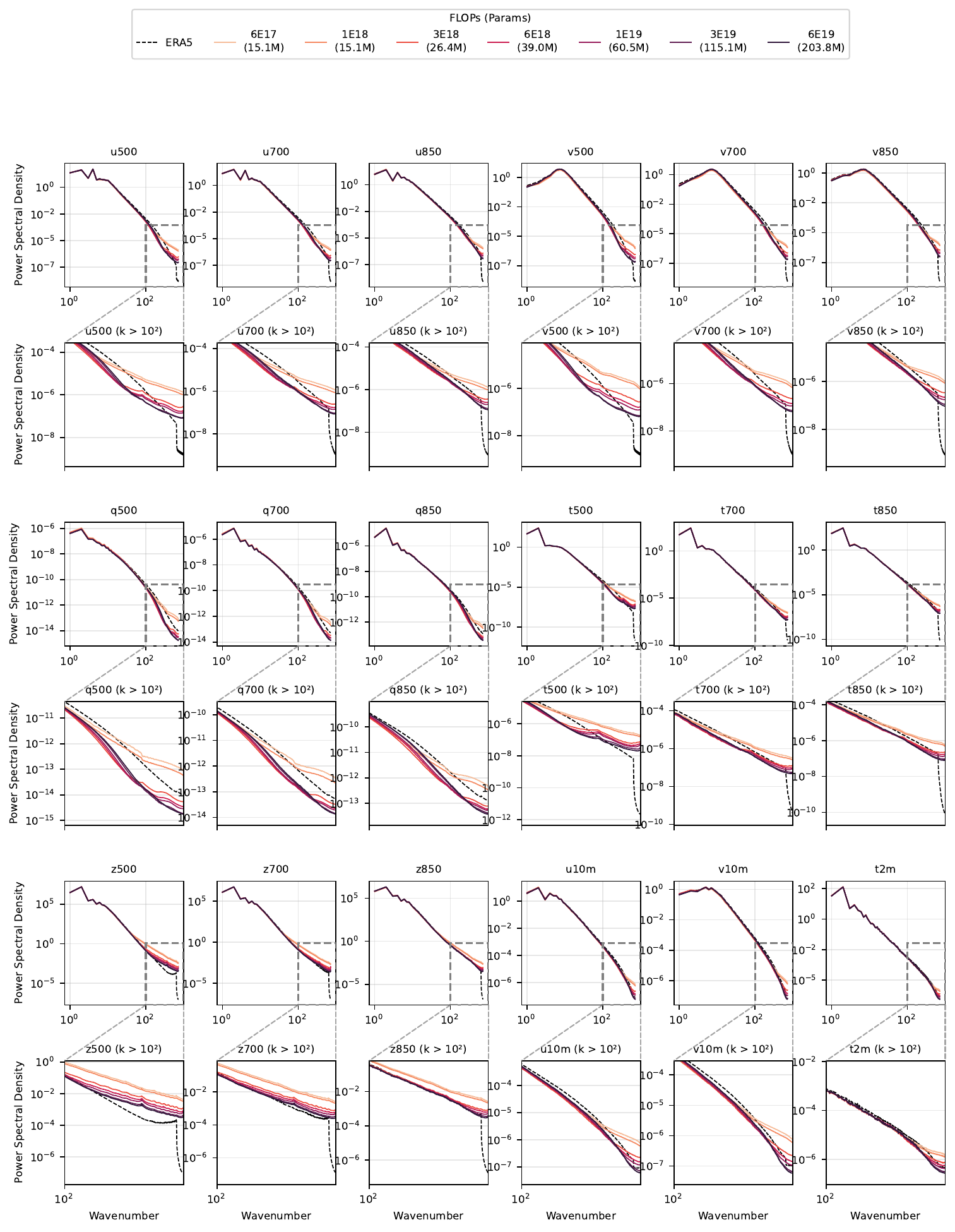}
    \caption{\textbf{PSD performance at lead time 120 hours of the closest compute-optimal {\swin} models at each compute budget for a larger subset of variables}. We show Fig. \ref{fig:spectra-flops-120h} for more variables.}
    \label{fig:spectra-flops-120h-all}
\end{figure}

\noindent \textbf{Results for more variables. }
We show the comparison with GraphCast and HRES for more variables in Fig. \ref{fig:rmse-all}, Fig. \ref{fig:spectra-all-24h}, and Fig. \ref{fig:spectra-all-120h}, where we see that the 8-step {\AR} {\swin} is similar in performance to GraphCast. We also show the performance with compute for all the additional variables in Fig. \ref{fig:rmse-flops-all} and Fig. \ref{fig:spectra-flops-120h-all}, with similar behavior as the variable subset.

\end{document}